\documentclass[lettersize,journal]{IEEEtran}

\usepackage{xcolor}
\usepackage{array}
\usepackage[caption=false,font=normalsize,labelfont=sf,textfont=sf]{subfig}
\usepackage{textcomp}
\usepackage{stfloats}
\usepackage{url}
\usepackage{verbatim}
\usepackage{graphicx}
\usepackage{cite}
\usepackage{algpseudocode,algorithm,algorithmicx}

\usepackage{tikz}
\usetikzlibrary{trees}
\usepackage{tabularx}

\usepackage{epstopdf}

\usepackage{amsmath,amssymb,amsfonts}
\usepackage{optidef}
\usepackage{textcomp}
\usepackage{multirow}
\usepackage{romannum}
\usepackage{diagbox}
\usepackage{enumitem}
\usepackage{bm}
\usepackage{amsthm}
\newtheorem*{remark}{Remark}

\usepackage{hyperref}

\begin{document}
%

\title{Recent Advances in Path Integral Control for Trajectory Optimization: An Overview in Theoretical and Algorithmic Perspectives}



\author{Muhammad Kazim, JunGee Hong, Min-Gyeom Kim and Kwang-Ki K. Kim${}^{*}$
\thanks{The authors are with the department of electrical and computer engineering at Inha University,
Incheon 22212, Republic of Korea.}
\thanks{${}^{*}$Corresponding author (Email: {\tt kwangki.kim@inha.ac.kr})}
\thanks{This research was supported by the Basic Science Research Program through the National Research Foundation of Korea (NRF) funded by the Ministry of Education (NRF-2022R1F1A1076260).}
}

%




\maketitle

\begin{abstract}
This paper presents a tutorial overview of path integral (PI) approaches for stochastic optimal control and trajectory optimization.
We concisely summarize the theoretical development of path integral control to compute a solution for stochastic optimal control and provide algorithmic descriptions of the cross-entropy (CE) method, an open-loop controller using the receding horizon scheme known as the model predictive path integral (MPPI), and a parameterized state feedback controller based on the path integral control theory. 
We discuss policy search methods based on path integral control, efficient and stable sampling strategies, extensions to multi-agent decision-making, and MPPI for the trajectory optimization on manifolds. 
For tutorial demonstrations, some PI-based controllers are implemented in Python, MATLAB and ROS2/Gazebo simulations for trajectory optimization. The simulation frameworks and source codes are publicly available at~\href{https://github.com/INHA-Autonomous-Systems-Laboratory-ASL/An-Overview-on-Recent-Advances-in-Path-Integral-Control}{\color{blue}the github page}.
\end{abstract}

\begin{IEEEkeywords}
Stochastic optimal control, trajectory optimization, Hamilton-Jacobi-Bellman equation, Feynman-Kac formula, path integral, variational inference, KL divergence, importance sampling, model predictive path integral control, policy search, policy improvement with path integrals, planning on manifolds.
\end{IEEEkeywords}


\section{Introduction}\label{sec:intro}


Trajectory optimization for motion or path planning~\cite{von1992direct,betts1998survey,rao2014trajectory} is a fundamental problem in autonomous systems~\cite{choset2005principles,latombe2012robot,lavalle2006planning}. Several requirements must be simultaneously considered for autonomous robot motion, path planning, navigation, and control. Examples include the specifications of mission objectives, examining the certifiable dynamical feasibility of a robot, ensuring collision avoidance, and considering the internal physical and communication constraints of autonomous robots.

In particular, generating an energy-efficient and collision-free safe trajectory is of the utmost importance during the process of autonomous vehicle driving~\cite{paden2016survey,claussmann2019review,teng2023motion}, autonomous racing drone~\cite{song2021autonomous,han2021fast,hanover2023autonomous}, unmanned aerial vehicles~\cite{lan2021survey}, electric vertical take-off and landing (eVTOL) urban air mobility (UAM)~\cite{wang2021trajectory,park2023trajectory,pradeep2020wind}, missile guidance~\cite{kwon2020convex,roh2020l1}, space vehicle control, and satellite attitude trajectory optimization~\cite{garcia2005trajectory,weiss2014spacecraft,gatherer2019magnetorquer,malyuta2021advances,dearing2022efficient}


From an algorithmic perspective, the complexity of motion planning is NP-complete~\cite{canny1988complexity}.
Various computational methods have been proposed for motion planning, including sampling-based methods~\cite{elbanhawi2014sampling,kingston2018sampling}, nonlinear programming (NLP)~\cite{betts2010practical,kelly2017introduction}, sequential convex programming (SCP)~\cite{bonalli2019gusto,howell2019altro,manyam2021trajectory,malyuta2022convex},
differential dynamic programming (DDP)~\cite{jacobson1970differential,mayne1973differential,xie2017differential,chen2019autonomous,pavlov2021interior,cao2022direct,chatzinikolaidis2021trajectory,kim2022extension}, 
hybrid methods~\cite{zhong2020hybrid}, and differential-flatness-based optimal control~\cite{sun2022comparative,faessler2017differential}. 

Optimization methods can explicitly perform safe and efficient trajectory generation for path and motion planning. The two most popular optimal path and motion planning methods for autonomous robots are gradient- and sampling-based methods for trajectory optimization. The former frequently assumes that the objective and constraint functions in a given planning problem are differentiable and can rapidly provide a locally optimal smooth trajectory~\cite{FORCESPro}.
However, numerical and algorithmic computations of derivatives (gradient, Jacobian, Hessian, etc.) are not stable in the worst case; preconditioning and prescaling must be accompanied by and integrated into the solvers. In addition, integrating the outcomes of perception obtained from exteroceptive sensors such as LiDARs and cameras into collision-avoidance trajectory optimization requires additional computational effort in dynamic and unstructured environments. 

Sampling-based methods for trajectory generation do not require function differentiability. Therefore, they are more constructive than the former gradient-based optimization methods for modelling obstacles without considering their shapes in the constrained optimization for collision-free path planning~\cite{kingston2018sampling,gammell2020asymptotically}. In addition, sampling-based methods naturally perform explorations, thereby avoiding the local optima. However, derivative-free-sampling-based methods generally produce coarse trajectories with zigzagging and jerking movements. For example, rapidly exploring random trees (RRT) and probabilistic roadmap (PRM) methods generate coarse trajectories~\cite{kuffner2000rrt,lavalle2006planning,elbanhawi2014sampling,janson2018deterministic}. To mitigate the drawbacks of gradient- and sampling-based methods while maintaining their advantages, a hybrid method that combines them can be considered, as proposed in~\cite{campos2017hybrid,ravankar2020hpprm,kiani2021adapted,yu2022novel}.

Several open-source off-the-shelf libraries are available for implementing motion planning and trajectory optimization, which include
Open Motion Planning Library (OMPL)~\cite{sucan2012open},
Stochastic Trajectory Optimization for Motion Planning (STOMP)~\cite{kalakrishnan2011stomp},
Search-Based Planning Library (SBPL)~\cite{likhachev2010search},
And Covariant Hamiltonian Optimization for Motion Planning (CHOMP)~\cite{zucker2013chomp}.

The path integral (PI) for stochastic optimal control, which was first presented in~\cite{kappen2005linear}, is another promising approach for sampling-based real-time optimal trajectory generation. In the path integral framework, the stochastic optimal control associated with trajectory optimization is transformed into a problem of evaluating a stochastic integral, for which Monte Carlo importance sampling methods are applied to approximate the integral. 
It is also closely related to the cross-entropy method~\cite{rubinstein2004cross}~for stochastic optimization and model-based reinforcement learning~\cite{gomez2014policy} for decision making.
The use of path integral control has recently become popular with advances in high-computational-capability embedded processing units~\cite{williams2017model} and efficient Monte Carlo simulation techniques~\cite{rubinstein2016simulation,thijssen2018consistent}. 


There are several variations of the PI control framework. The most widely used method in robotics and control applications is the model predictive path integral (MPPI) control, which provides a derivative-free sampling-based framework to solve finite-horizon constrained optimal control problems using predictive model-based random rollouts in path-integral approximations~\cite{williams2016aggressive,williams2017model,williams2018information,williams2019mppi}. However, despite its popularity, the performance and robustness of the MPPI are degraded in the presence of uncertainties in the simulated predictive model, similar to any other model-based optimal control method. To take the plant-model mismatches of simulated roll-outs into account~\cite{pan2015sample,abraham2020model}, adaptive MPPI~\cite{pravitra2020}, learning-based MPPI~\cite{williams2015gpu,okada2017path,mohamed2023gp} and tube-based MPPI~\cite{gandhi2021robust}, uncertainty-averse MPPI~\cite{arruda2017uncertainty}, fault-tolerant MPPI~\cite{raisi2022fault}, safety-critical MPPI using control barrier function (CBF)~\cite{zeng2021safety,tao2022path,yin2023shield}, and covariance steering MPPI~\cite{yin2022trajectory} have been proposed.
Risk-aware MPPI based on the conditional value-at-risk (CVaR) have also been investigated for motion planning with probabilistic estimation uncertainty in partially known environments~\cite{barbosa2021risk,wang2021adaptive,cai2022probabilistic,yin2023risk}.

The path integral (PI) control framework can also be combined with parametric and nonparametric policy search methods and improvements.
For example, RRT is used to guide PI control for exploration~\cite{tao2023rrt} and PI control is used to guide policy searches for open-loop control~\cite{theodorou2012relative}, parameterized movement primitives~\cite{ijspeert2002learning,theodorou2010generalized}, and feedback control~\cite{levine2013guided,gomez2014policy,montgomery2016guided}.
To smoothen the sampled trajectories generated from the PI control, gradient-based optimization, such as DDP, is combined with MPPI~\cite{kim2022mppi}, regularized policy optimization based on the cross-entropy method is used~\cite{thalmeier2020adaptive}, and 
it is also suggested to smooth the resulting control sequences using a sliding window~\cite{sarkka2008unscented,ruiz2017particle} and a Savitzky-Golay filter (SGF)~\cite{williams2018information,neve2022comparative} and introducing an additional penalty term corresponding to the time derivative of the action~\cite{kim2022smooth}.

PI control is related to several other optimal control and reinforcement learning strategies. 
For example, variational methods of path integral optimal control are closely related to entropy-regularized optimal control~\cite{lefebvre2022entropy,lambert2020stein} and maximum entropy RL (MaxEnt RL)~\cite{theodorou2010generalized,eysenbach2019if}.
The path integral can be considered as a probabilistic inference for stochastic optimal control~\cite{whittle1991likelihood,kappen2012optimal,watson2020stochastic} and reinforcement learning~\cite{haarnoja2017reinforcement,levine2018reinforcement}.

One of the most important technical issues in the practical application of path integral control is the sampling efficiency. Various importance sampling strategies have been suggested for rollouts in predictive optimal control. Several importance sampling (IS) algorithms exist ~\cite{martino2015adaptive,stich2017safe} with different performance costs and benefits, as surveyed in~\cite{bugallo2017adaptive}. Adaptive importance sampling (AIS) procedures are considered within the optimal control~\cite{kappen2016adaptive,thijssen2018consistent} and MPPI control~\cite{asmar2023model}. In addition to the AIS algorithms, learned importance sampling~\cite{carius2022constrained}, general Monte Carlo methods~\cite{Arouna+2004+1+24,rubinstein2016simulation}, and cross-entropy methods ~\cite{de2005tutorial,kobilarov2012cross,zhang2014applications} have been applied to PI-based stochastic optimal control.

Various case studies of PI-based optimal control have been published~\cite{testouri2023towards}: autonomous driving~\cite{williams2016aggressive,gandhi2021robust,mohamed2022autonomous,ha2019topology}, autonomous flying~\cite{mohamed2020model,pravitra2021flying,houghton2022path,higgins2023model}, space robotics~\cite{raisi2022fault}, autonomous underwater vehicles~\cite{nicolay2023enhancing}, and robotic manipulation planning~\cite{hou2022robotic,yamamoto2020path}. Path integral strategies for optimal predictive control have also been adopted to visual servoing techniques~\cite{mohamed2022autonomous,mohamed2021sampling,mohamed2021mppi,costanzo2023modeling}.
Recently, the MPPI was integrated into Robot Operating Systems 2 (ROS 2)~\cite{macenski2023desks}, an open-source production-grade robotics middleware framework~\cite{macenski2022robot}.

We expect that more applications of path integral control will emerge, particularly with a focus on trajectory optimization of motion planning for autonomous systems such as mobile robots, autonomous vehicles, drones, and service robots. In addition, it has been shown that path integral control can be easily extended to the cooperative control of multi-agent systems~\cite{van2008graphical,thijssen2016path,gomez2016real,wan2021cooperative,varnai2022multi}.

There are still issues that must be addressed for scalable learning with safety guarantees in path integral control and its extended variations. 
\begin{itemize}
\item
Exploration-exploitation tradeoff is still not trivial,
\item
Comparisons of MPC-like open-loop control and parameterized state-feedback control should be further investigated as problem and task-specific policy parameterization itself is not trivial,
\item
Extensions to output-feedback and dual control have not yet been studied, and 
\item
Extensions to cooperative and competitive multi-agent trajectory optimization with limited inter-agent measurements and information should be further investigated.  
\end{itemize}

The remainder of this paper is organized as follows:
Section~\ref{sec:PI:overview} presents the overview of some path integral control methods.
Section~\ref{sec:PI:theory} reviews the theoretical background of path integral control and its variations. 
Section~\ref{sec:PI:alg} describes the algorithmic implementation of several optimal control methods that employ a path-integral control framework.
In Section~\ref{sec:sim:matlab}, two MATLAB simulation case studies are presented to demonstrate the effectiveness of predictive path integral control. 
Section~\ref{sec:sim:gazebo} presents the four ROS2/Gazebo simulation results of trajectory optimization for autonomous mobile robots, in which MPPI-based local trajectory optimization methods are demonstrated for indoor and outdoor robot navigation.
In Section~\ref{sec:discussion}, extensions of path integral control to policy search and learning, improving sampling efficiency, multi-agent decision making, and trajectory optimization of manifolds are discussed.  
Section~\ref{sec:conclusion} concludes the paper and suggests directions for future research and development of path-integral control, especially for autonomous mobile robots.


\section{Overview of path integral controllers}\label{sec:PI:overview}

\begin{figure}[t]
	\centering
	\includegraphics[width=.99\linewidth]{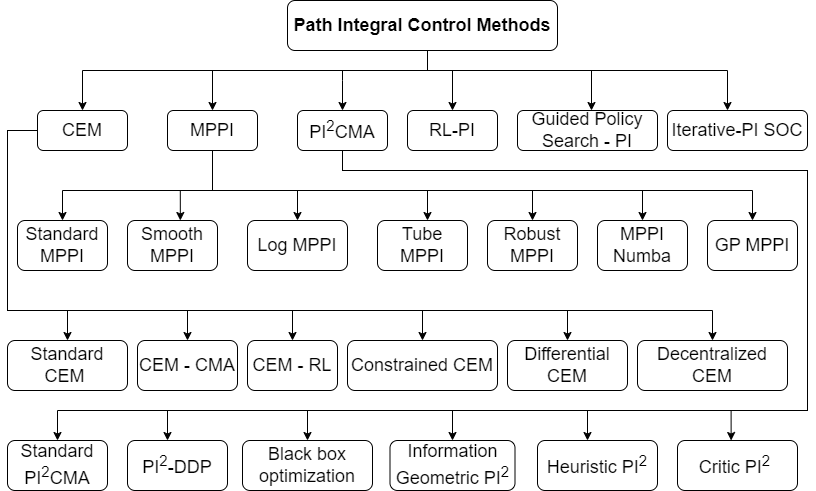}\vspace{-2mm}
	\caption{Hierarchical classification of various path integral control methods.}\vspace{3mm}
	\label{fig:pathintegral_architecture}
\end{figure}

\begin{table}[!t]
\begin{tabularx}{\linewidth}{|>{\hsize=.2\hsize}X|>{\hsize=.73\hsize}X|>{\hsize=.07\hsize}X|}
\hline
\textbf{Type} & \textbf{Description} & \textbf{Ref.} \\
\hline
{\vspace*{1pt} Standard CEM} & The standard CEM is a unified probabilistic approach for tackling combinatorial optimization, Monte-Carlo simulation, and machine learning challenges. & {\vspace*{0pt} \cite{rubinstein2004cross}} \\
\hline
{\vspace*{2pt} CEM-RL} & CEM-RL merges CEM and TD3 to enhance DRL, optimizing both performance and sample efficiency. & { \cite{pourchot2018cem}} \\
\hline
{\vspace*{2pt} Constrained CEM} & Constrained CEM is a safety-focused reinforcement learning method that learns feasible policies while effectively tracking and satisfying constraints. & {\vspace*{0pt} \cite{wen2018constrained}} \\
\hline
{\vspace*{0pt} Differentiable CEM} & This introduces a gradient-based optimization into the CEM framework to improve the convergence rate. & {\cite{amos2020differentiable} }   \\
\hline
{\vspace*{0pt} Decentralized CEM} & Decentralized CEM improves Cross-Entropy Method efficiency by employing a distributed ensemble of CEM instances to mitigate local optima traps. & {\cite{zhang2022simple}} \\
\hline
\end{tabularx}\vspace{2mm}
\caption{Types of CEM algorithms and their description.}
\label{tab:cem_algorithms}
\vspace{2mm}
\begin{tabularx}{\linewidth}{|>{\hsize=.2\hsize}X|>{\hsize=.73\hsize}X|>{\hsize=.07\hsize}X|}
\hline
\textbf{Type} & \textbf{Description} & \textbf{Ref.} \\
\hline
{\vspace*{0pt} Standard MPPI} & MPPI leverages parallel optimization and generalized importance sampling for efficient navigation in nonlinear systems. & {\cite{williams2017model}}  \\
\hline
{\vspace*{0pt} Robust MPPI} & It enhances off-road navigation with an augmented state space and tailored controllers, offering improved performance and robustness. & {\cite{gandhi2021robust}}  \\
\hline
{\vspace*{0pt} Smooth MPPI} & It enhances MPPI by incorporating input-lifting and a novel action cost to effectively reduce command chattering in nonlinear control tasks. & { \cite{kim2022smooth}} \\
\hline
{\vspace*{4pt} Log MPPI} & It enhances robotic navigation by using a normal log-normal mixture for efficient and feasible trajectory sampling in complex environments. & { \cite{mohamed2022autonomous}} \\
\hline
{\vspace*{7pt} Tube MPPI} & Tube MPPI merges MPPI with Constrained Covariance Steering for robust, constraint-efficient trajectory optimization in stochastic systems. & { \cite{balci2022constrained}}  \\
\hline
{\vspace*{3pt} MPPI Numba} & This research develops a neural network-based, risk-aware method for off-road navigation, enhancing success and efficiency by analyzing empirical traction distributions. & {\vspace*{0pt} \cite{cai2022probabilistic}} \\
\hline
{\vspace*{12pt} GP-MPPI} & GP-MPPI combines MPPI with sparse Gaussian Processes for improved autonomous navigation in complex environments by learning model uncertainties and disturbances, and optimizing local paths. & {\vspace*{2pt}\cite{mohamed2023gp}}  \\
\hline
\end{tabularx}\vspace{2mm}
\caption{Types of MPPI algorithms and their description.}
\label{tab:mppi_algorithms}
\end{table}

\begin{table}[!t]
\begin{tabularx}{\linewidth}{|>{\hsize=.2\hsize}X|>{\hsize=.73\hsize}X|>{\hsize=.07\hsize}X|}
\hline
\textbf{Type} & \textbf{Description} & \textbf{Ref.} \\
\hline
{\vspace*{0pt} Standard \( \text{PI}^2\text{-CMA} \)} & It is an algorithm that merges PI$^2$-CMA to optimize policies and auto-tune exploration noise in reinforcement learning. & {\cite{stulp2012path}} \\
\hline
{\vspace*{0pt} Black-box Optimization} & It examines the convergence of RL and black-box optimization in PI, introducing the efficient PIBB algorithm.  & { \cite{stulp2012policy}}  \\
\hline
{\vspace*{2pt} PI$^2$-DDP} & It is an advanced RL method that combines the gradient extraction principles of PI$^2$-CMA with the feedback mechanism of DDP. & { \cite{lefebvre2019path}}  \\
\hline
{\vspace*{0pt} Information Geometric PI$^2$} & It reexamines PI$^2$, linking it with evolution strategies and information-geometric optimization to refine its cost minimization approach using natural gradient descent. & {\vspace*{0pt} \cite{varnai2020path}}  \\
\hline
{\vspace*{15pt} Heuristic PI$^2$} & It introduces PI2-CMA-KCCA, a RL algorithm that accelerates robot motor skill acquisition by using heuristic information from Kernel Canonical Correlation Analysis and CMA to guide Path Integral Policy Improvement. & {\vspace*{4pt} \cite{fu2020compound}} \\
\hline
{\vspace*{8pt} Critic PI$^2$} & Critic PI$^2$ combine trajectory optimization and deep actor-critic learning in a model-based RL framework for improved efficiency in continuous control tasks. & {\vspace*{0pt} \cite{ba2021critic}} \\
\hline
\end{tabularx}\vspace{2mm}
\caption{Types of \( \text{PI}^2\text{-CMA} \) algorithms and their description.}
\label{tab:pi2cma_algorithms}
\end{table}

Path integral (PI) control methods stand at the forefront of contemporary research in stochastic optimal control and trajectory optimization. These techniques, primarily characterized by their robustness and versatility, have been developed to tackle complex control tasks under uncertainty. Central to these methods are Cross-Entropy Method (CEM), Model Predictive Path Integral (MPPI), and PI$^{2}$-CMA, each with distinct variations and algorithmic developments. In Fig. \ref{fig:pathintegral_architecture}, we mentioned some hierarchical structure of path integral control methods. Below, we present a structured overview of these methods, outlining their types and key characteristics.

The Cross-Entropy Method is a probabilistic technique that iteratively refines control policies by minimizing a cost function and employing importance sampling. It has evolved significantly since its inception. Some of the CEM algorithms and there key characteristics are given in Table~\ref{tab:cem_algorithms}.
MPPI is an open-loop controller using the receding horizon scheme. It has been pivotal for real-time control with various developments focusing on improving efficiency and reducing computational load. Some of the MPPI algorithms and there key characteristics are given in Table~\ref{tab:mppi_algorithms}.
PI$^2$-CMA represents an amalgamation of path integral control with Covariance Matrix Adaptation. It's especially suitable for problems where the cost landscape is highly non-convex or unknown. Some of the MPPI algorithms and there key characteristics are given in Table~\ref{tab:pi2cma_algorithms}.

\section{Path Integral Control: Theory}\label{sec:PI:theory}
\subsection{Stochastic optimal control}
\label{sec:PI:theory:1}
\noindent
\textbf{Controlled stochastic differential equation}
Consider a controlled stochastic differential equation of the following form:
\begin{equation}\label{eq:sde:general}
dX_{t} = f(t,\!X_{t},\! \pi(t,X_{t})) dt + g(t,\!X_{t}, \!\pi(t,X_{t})) dW_{t},
\end{equation}
where the initial condition is given by $X_{0} = x_0 \in {\mathbb R}^{n}$, and $W_{t}$ is a standard Brownian motion.
The solution to the SDE~\eqref{eq:sde:general} associated with the Markov policy $\pi: {\mathbb R} \times {\mathbb R}^{n} \rightarrow {\mathbb R}^{m}$ is denoted by $X_{t}^{\pi}$.

\noindent
\textbf{Cost-to-go function}
For a given Markov policy $\pi$, the cost-to-go corresponding to policy $\pi$ is defined as
\begin{equation}
G_{t}^\pi = \phi(X_{T}^\pi) + \int_{t}^{T} \!\! L(s, X_{s}^{\pi}, \pi(s, X_{s}^{\pi})) ds ,
\end{equation}
where $T>0$ denotes the terminal time, $L: {\mathbb R} \times {\mathbb R}^{n} \times {\mathbb R}^{m} \rightarrow {\mathbb R}$ denotes the running cost, and $\phi:{\mathbb R}^{n} \rightarrow {\mathbb R}$ denotes the terminal cost. 

\noindent
\textbf{Expected and optimal cost-to-go function}
The expected cost-to-go function is defined as
\begin{equation}
V^{\pi}(t,x) = {\mathbb E} [G_{t}^\pi  | X_{t}^{\pi} = x]
\end{equation}
where the expectation is considered with respect to the probability of the solution trajectory for the SDE~\eqref{eq:sde:general} with an initial time and condition $(t,x)\in{\mathbb R}\times {\mathbb R}^n$. The goal of the stochastic optimal control is to determine the optimal policy. 
\begin{equation}
\pi^* = \arg\min_{\pi} V^{\pi}(t,x) 
\end{equation}
The corresponding optimal cost-to-go function is defined as
\begin{equation}
V^{*}(t,x) = V^{\pi^*}({t}, x) = \min_{\pi} V^{\pi}({t},x)
\end{equation}
for each $(t,x) \in [0,T] \times {\mathbb R}^{n}$,

\subsection{The Hamilton-Jacobi-Bellman equation}
\label{sec:PI:theory:2}
The Hamilton-Jacobi-Bellman equation for the optimal cost-to-go function is defined as follows ~\cite{fleming2006controlled}
\begin{equation}\label{eq:HJB}
\min_{\pi} \left\{ L(t,x,\pi(t,x)) + {\mathcal T}^\pi V^{*}\!(t,x) \right \} = 0
\end{equation}
$(t,x) \in [0,T] \times {\mathbb R}^{n}$.
where
\begin{equation}
\begin{split}
& {\mathcal T}^\pi V^{*}(t,x) \\
& \,  =\! \lim_{h\rightarrow0+}\!\!\left( {\mathbb E}[V^*({t\!+\!h}, X_{t+h}^\pi) | X_{t}^\pi \!= x)] - V^*\!({t},x) \right)
\end{split}
\end{equation}
is a backward evolution operator defined on the functions of the class ${\mathcal C}^{1}\!\times{\mathcal C}^{2}$.
Additionally, the boundary condition is given by $V^{*}(T,x) = \phi(x)$.

\subsection{Linearization of the HJB PDE}
\label{sec:PI:theory:3}
\noindent
\textbf{Control affine form and a quadratic cost}
As a special case of \eqref{eq:sde:general}, we consider the controlled dynamics (diffusion process)
\begin{equation}\label{eq:sde:affine}
dX_{t} = f(t,\!X_{t}) dt + g(t,\!X_{t}) \!\left( \pi(t,X_{t})) dt+ dW_{t} \right),
\end{equation}
and the cost-to-go
\begin{equation}\label{eq:cost:quad}
\begin{split}
G_{t}^\pi = \ 
&\phi(X_{T}^\pi) + \int_{t}^{T}\pi(s,\!X_{s}^\pi)^{\!\top} \! dW_{\!s}  \\
&+ \int_{t}^{T} \!\!\!\left(\! q(s,\!X_{s}^\pi) + \frac{1}{2} \pi(s,\!X_{s}^\pi)^{\!\top} \!\pi(s,\!X_{s}^\pi) \!\right) \!ds
\end{split}
\end{equation}
where $X_{s}^\pi$ is the solution to the SDE~\eqref{eq:sde:affine} for $s\in[t,T]$ with the initial condition $X_{t}^{\pi}=x_t$.
In this study, we assume $G_{t}^\pi > 0$ for all $t \in [0,T]$ and any (control) policy $\pi$.

\begin{remark}
{\it
Notice that
\begin{itemize}
	\item[]
	1) $G_{t}^{\pi}$ is not adaptive with respect to the Brownian motion as it depends on $(X_{\tau}^\pi)$ for $\tau>t$.
	\item[]
	2) The second term in~\eqref{eq:cost:quad} is a stochastic integral with respect to the Brownian motion and it vanishes when taking expectation.
	This term will play an essential role when applying a change of measure. \hfill$\square$
\end{itemize}
}
\end{remark}

The goal of stochastic optimal control for the dynamics~\eqref{eq:sde:affine}~and~the cost-to-go~\eqref{eq:cost:quad} is to determine an optimal policy
that minimizes the expected cost-to-go with respect to the policy.
\begin{align}
V^{*}(t,x) &: = \min_{\pi} {\mathbb E} [G_{t}^\pi  | X_{t}^{\pi} = x], \label{eq:soc:valuefunc} \\
\pi^*(t,x) &: = \arg\min_{\pi} V^{\pi}(t,x) ,  \label{eq:soc:optcontr} 
\end{align}
where the expectation ${\mathbb E}$ is considered with respect to the stochastic process $X_{t:T}^{\pi} \sim {\mathcal P}^{\pi}$ which is the (solution) path of SDE~\eqref{eq:sde:affine}.

\noindent
\textbf{Theorem 1:}~\cite{fleming2006controlled,oksendal2013stochastic,fleming2012deterministic} The solution of the stochastic optimal control in \eqref{eq:soc:valuefunc} and \eqref{eq:soc:optcontr} is given as
\begin{equation}\label{eq:thm1:1}
V^{*}(t,x) = - \ln {\mathbb E}\!\left[ e^{-G_{t}^{\pi}} | X_{t}^{\pi} = x\right] , \\
\end{equation}
and
\begin{equation}\label{eq:thm1:2}
\begin{split}
\pi^* (t, x)    
&= \pi (t, x) \\
& \quad + \lim_{s\rightarrow t+} \! \frac{{\mathbb E}  [(W_{s} \!-\! W_{t}) e^{-G_{t}^{\pi}}\!| X_{t}^{\pi} = x]}{{\mathbb E}  [({s} \!-\! {t}) e^{-G_{t}^{\pi}} \!| X_{t}^{\pi} = x]} ,
\end{split}
\end{equation}
where $\pi(t,x)$ denotes an arbitrary Markov policy. \hfill$\blacksquare$ 

Because the solution represented in Theorem 1 is defined in terms of a path integral for which the expectation ${\mathbb E}$ is taken with respect to the random process $X_{t:T}^{\pi} \sim {\mathcal P}^{\pi}$, that is, the (solution) path of the SDE~\eqref{eq:sde:affine}, this class of stochastic optimal control with control-affine dynamics and quadratic control costs is called the path integral (PI) control. 

\noindent
\textbf{Solution of the HJB equation}
For stochastic optimal control of the dynamics~\eqref{eq:sde:affine}~and~\eqref{eq:cost:quad}, the HJB equation can be rewritten as~\cite{oksendal2013stochastic}  
\begin{equation}\label{eq:HJB:special}
\begin{split}
\min_{u} \left\{ q + \frac{1}{2}u^\top u + {\mathcal T}^{u} V^{*} \right \} = 0
\end{split}
\end{equation}
where 
\begin{equation}
{\mathcal T}^{u} V^{*}
=
\partial_{t} V^* + (f+gu)^\top \partial_{x} V^* + \frac{1}{2} \textrm{Tr}(gg^{\top}\partial_{xx} V^*)
\end{equation}
with the boundary condition $V^*(T,x) = \phi(x)$. In addition, the optimal state-feedback controller is given by
\begin{equation}\label{eq:HJB:special:sol}
u^*(t,x) = - g(t,x)^\top \partial_{x} V^*(t,x)\,.
\end{equation}
Here, Markov policy $\pi$ is replaced by state-feedback control $u$ without loss of generality.
The value (i.e., the optimal expected cost-to-go) function $V^*$ is defined as a solution to the second-order PDE~\cite{oksendal2013stochastic,fleming2006controlled}
\begin{equation}
\begin{split}
&0= q+ \partial_{t} V^* \!- \frac{1}{2} ( \partial_{x} V^* )^{\!\top} \!gg^\top\! \partial_{x} V^* + f^\top\! \partial_{x} V^*  \\
& \quad \ \ + \frac{1}{2} \textrm{Tr}(gg^{\top}\partial_{xx} V^*) \,. 
\end{split}
\end{equation}

\noindent
\textbf{Linearization via exponential transformation}
We define an exponential transformation as follows:
\begin{equation}
\psi(t,x) = \exp\! \left(-\frac{1}{\lambda}V^*(t,x) \right)
\end{equation}
that also belongs to class ${\mathcal C}^{1}\!\times {\mathcal C}^{2}$ provided $V^*(t,x)$ does.
Applying the backward evolution operator of the \emph{uncontrolled} process, that is, $u=0$, to the function $\psi(t,x)$, we obtain
\begin{equation}\label{eq:pde:CK}
{\mathcal T}^{0}\psi
= \partial_{t} \psi  + f^\top \partial_{x} \psi + \frac{1}{2} \textrm{Tr}(gg^{\top}\partial_{xx} \psi) 
= \frac{1}{\lambda} q \psi
\end{equation}
which is a linear PDE with the boundary condition $\psi(T,x) = \exp(-\phi(x)/\lambda)$. This linear PDE is known as the backward Chapman-Kolmogorov PDE~\cite{oksendal2013stochastic}.

\subsection{The Feynman-Kac formula}
\label{sec:PI:theory:4}
The Feynman-Kac lemma~\cite{oksendal2013stochastic}~provides a solution to the linear PDE~\eqref{eq:pde:CK}
\begin{equation}
\psi(t,x) 
= {\mathbb E}\! \left[\exp\!\left(\!\left(\! -\frac{1}{\lambda} \! \int_{t}^{T}\!\! q(s,X_{s}^{0})ds \!\right) \!\psi(T,x_{T})\! \right)\! \right]
\end{equation}
where $\psi(T,x_{T}) = \exp(-\phi(x)/\lambda)$. In other words, 
\begin{equation}
\psi(t,x) = {\mathbb E}\! \left[ \exp\!\left(\! -\frac{1}{\lambda}G_{t}^{0} \!\right) \left| X_{t}^{0} = x \right. \right]
\end{equation}
where the expectation ${\mathbb E}$ is taken with respect to the random process $X_{t:T}^{0} \sim {\mathcal P}^{0}$; that is, the (solution) path of the uncontrolled version of the SDE~\eqref{eq:sde:affine}
\begin{equation}\label{eq:sde:affine:uncontrolled}
dX_{t}^{0} = f(t,\!X_{t}^{0}) dt + g(t,\!X_{t}^{0}) dW_{t}
\end{equation}
and the uncontrolled cost-to-go is given by
\begin{equation}\label{eq:cost:uncontrolled}
G_{t}^{0} = \phi(x_{T}^{0}) + \int_{t}^{T} \!\! q(s,X_{s}^{0}) ds 
\end{equation}
which is again a nonadaptive random process. From the definition of $\psi$, this gives us a path-integral form for the value function:
\begin{equation}\label{eq:FK:value}
V^*(t,x) = -\lambda \ln {\mathbb E}\! \left[ \exp\!\left(\! -\frac{1}{\lambda}G_{t}^{0} \!\right) \!\left| X_{t}^{0} = x \right. \right]
\end{equation}
$(t,x) \in [0,T] \times {\mathbb R}^{n}$.

\subsection{Path integral for stochastic optimal control}
\label{sec:PI:theory:5}
\noindent
\textbf{Path integral control}
Although the Feynman-Kac formula presented in Section~\ref{sec:PI:theory:4}~provides a method to compute or approximate the value function~\eqref{eq:FK:value}, it is still not trivial to obtain an optimal Markov policy because the optimal controller in~\eqref{eq:HJB:special:sol} is defined in terms of the gradient of $V^*$, not by $V^*$. From~\eqref{eq:HJB:special:sol}~and~Theorem 1, combined with the path integral control theory~\cite{thijssen2016path,kappen2007introduction,kappen2011optimal,theodorou2011iterative,thijssen2015path,williams2019mppi}, we have
\begin{equation}\label{eq:picontrol}
\begin{split}
 u^*(t,x) &= - g(t,x)^{\top} \partial_{x} V^*(t,x) \\
          & = g(t,x)^{\top} \partial_{x} \ln \psi(t,x) \\
          &  = g(t,x)^{\top} \lim_{s\rightarrow t+}  \frac{{\mathbb E}_{{\mathcal P}^{0}} [ \exp\left(-\frac{1}{\lambda}G_{t}^{0} \right)\int_{t}^{s} g(\tau,X_{\tau})dW_{\tau}]}{(s-t){\mathbb E}_{{\mathcal P}^{0}}\![\exp\left( -\frac{1}{\lambda}G_{t}^{0} \right)]}
\end{split}
\end{equation}
where the initial condition is $X_{t}^{0} = x$. This is equivalent to~\eqref{eq:thm1:2}~in Theorem 1.

\noindent
\textbf{Information theoretic stochastic optimal control}
Regularized cost-to-go function 
\begin{equation}
{\mathcal S}_{t}({\mathcal P}^{\pi}) = G_{t}^{0} + \lambda \ln\frac{d{\mathcal P}^{\pi}}{d{\mathcal P}^{0}}
\end{equation}
where $G_{t}^{0}$ is the state-dependent cost given in~\eqref{eq:cost:uncontrolled} and $\frac{d{\mathcal P}^{\pi}}{d{\mathcal P}^{0}}$ is the Radon-Nikodym derivative\footnote{This R-N derivative $\frac{d{\mathcal P}^{\pi}}{d{\mathcal P}^{0}}$ denotes the density of ${\mathcal P}^{\pi}$ relative to ${\mathcal P}^{0}$. We assume that ${\mathcal P}^{\pi}$ is absolutely continuous with respect to ${\mathcal P}^{0}$, denoted by ${\mathcal P}^{\pi} \ll {\mathcal P}^{0}$.} for the probability measures ${\mathcal P}^{\pi}$ and ${\mathcal P}^{0}$. Total expected cost function 
\begin{equation}
\begin{split}
{\mathcal V}_{t}({\mathcal P}^{\pi}) 
&= {\mathbb E}_{{\mathcal P}^\pi}[{\mathcal S}_{t}({\mathcal P}^{\pi})] \\
&= {\mathbb E}_{P^\pi}[G_{t}^{0}] + \lambda D_{\rm KL}({\mathcal P}^{\pi}\|{\mathcal P}^{0})
\end{split}
\end{equation}
is known as the free energy of a stochastic control system~\cite{fleming2006controlled,fleming1995risk,theodorou2012relative,theodorou2015nonlinear}. There is an additional cost term for the KL divergence between ${\mathcal P}^{\pi}$ and ${\mathcal P}^{0}$ which we can interpret as a control cost.
From Girsanov’s theorem~\cite{liptser1977statistics,liptser2013statistics}, we obtain the following expression for the Radon-Nikodym derivative corresponding to the trajectories of the control-affine SDE~\eqref{eq:sde:affine}.
\begin{equation}\label{eq:RNd:1}
\frac{d{\mathcal P}^{\pi}}{d{\mathcal P}^{0}} =  \exp\! \left( \int_{t}^{T} \! \frac{1}{2} \|u_{s}\|^2 ds + u_{s}^\top \! dW_{s}  \right)
\end{equation}
where $u_{s} = \pi(s,X_{s}^{\pi})$ is the control input and the initial condition $X_{t}^{\pi} = X_{t}^{0} = x_{t}$ with an initial time $t \in [0,T]$ can be arbitrary.

The goal of KL control is to determine a probability measure that minimizes the expected total cost
\begin{equation}
{\mathcal P}^* 
= {\mathcal P}^{\pi^*} \!\!
= \arg \min_{{\pi}} {\mathcal V}_{t}({\mathcal P}^{\pi}) 
= \arg \min_{{\mathcal P} \in {\bm\Delta}^{\!\pi}} {\mathcal V}_{t}({\mathcal P})
 \,,
\end{equation}
provided ${\mathcal V}_{t}^* = \inf_{{\mathcal P}\in{\bm\Delta}^{\!\pi}} {\mathcal V}_{t}({\mathcal P})$ exists, where ${\bm\Delta}^{\!\pi}$ denotes the space of probability measures corresponding to a policy $\pi$.

\noindent
\textbf{Theorem 2:}~\cite{thijssen2016path,thijssen2015path}
The optimal regularized cost-to-go has zero variance and is the same as the expected total cost
\begin{equation}
{\mathcal S}_{t}({\mathcal P}^{*}) = {\mathcal V}_{t}({\mathcal P}^{*}) = - \lambda \ln {\mathbb E}_{{\mathcal P}^{0}}[\exp( - {G_{t}^{0}}/\lambda )] \,.
\end{equation}
which is equivalent to~\eqref{eq:FK:value} given in Section~\ref{sec:PI:theory:4}.
\hfill$\blacksquare$

In addition, the Radon-Nikodym derivative is given by
\begin{equation}\label{eq:RNd:2}
\frac{d{\mathcal P}^{*}}{d{\mathcal P}^{0}} = \frac{\exp({-G_{t}^{0}}/\lambda)}{{\mathbb E}_{{\mathcal P}^{0}}[\exp({-G_{t}^{0}}/\lambda)]}
\end{equation}
and combining~\eqref{eq:RNd:2}~with the R-N derivative~\eqref{eq:RNd:1}, we obtain
\begin{equation}
\begin{split}
\frac{d{\mathcal P}^{*}}{d{\mathcal P}^{\pi}} 
&= \omega_{t}^{\pi} \\
&= \exp\! \left(\!-\! \!\int_{t}^{T} \!\!\frac{1}{2} \|u_{s}\|^2 ds - u_{s}^\top \! dW_{s} - \frac{1}{\lambda} G_{t}^{0} \right)
\end{split}
\end{equation}
where $u_{s} = \pi(s,X_{s})$ is the control input following the policy $\pi:{\mathbb R}\times{\mathbb R}^{n}\rightarrow{\mathbb R}^{m}$, and $\omega_{t}^{\pi}$ is known as the importance weight~\cite{thijssen2015path,kappen2016adaptive}~that is also a random process. 


\section{Path Integral Control: Algorithms}\label{sec:PI:alg}

\subsection{MC integration: Model-based rollout}
\label{sec:PI:alg:mc}
Let a tuple $(\Omega, {\mathcal F}, {\mathcal Q})$ be a probability space with a random variable $X$ and consider the function $\ell({\bm X}) = \int_{0}^{T} L({\bm X}_{t}) dt$ or $\ell(X) = L({\bm X}_{T})$.
The main idea of path integral control is to compute the expectation
\begin{equation}
\rho = {\mathbb E}_{\mathcal Q}[\ell({\bm X})]
\end{equation}
using sampling-based methods, such as Monte Carlo simulations. In principle, function $\ell:\Omega \rightarrow {\mathbb R}$ can be any arbitrary real-valued function. A well-known drawback of Monte Carlo (MC) integration is its high variance.

\noindent
\textbf{Importance sampling} The goal of importance sampling~\cite{Arouna+2004+1+24,rubinstein2016simulation}~in MC techniques is to minimize the variance in the MC estimation of integration, ${\mathbb E}_{\mathcal Q}[\ell({\bm X})]={\mathbb E}_{\mathcal P}[\ell({\bm X})\frac{d{\mathcal Q}}{d{\mathcal P}}]$.
To reduce the variance, we want to find a probability measure ${\mathcal P}$ on $(\Omega, {\mathcal F})$ with which an unbiased MC estimate for $\rho$ is given by
\begin{equation}
\hat{\rho}({\mathcal P}) = \frac{1}{N_{\!s}} \! \sum_{i=1}^{N_{\!s}} \ell({\bm X}_{i})\frac{d{\mathcal Q}}{d{\mathcal P}}({\bm X}_{i}), 
\end{equation}
where the $i$th sampled path ${\bm X}_{i}$ is generated from density ${\mathcal P}$, denoted by ${\bm X}_{i}\sim{\mathcal P}$, for $i=1,\ldots, N_{\!s}$.

\noindent
\textbf{Multiple importance sampling} 
Multiple-based probability measures can also be used for the MC estimation.
\begin{equation}
\hat{\rho}(\{{\mathcal P}^{j}\}_{j=1}^{N_{\!p}}) \!=\! \frac{1}{N} \! \sum_{j=1}^{N_{\!p}} \! \sum_{i=1}^{N_{\!s}^{j}} \!\ell({\bm X}_{i}^{j})\frac{d{\mathcal Q}}{d{\mathcal P}^{j}}({\bm X}_{i}^{j}) \gamma^{j}\!({\bm X}_{i}^{j})
\end{equation}
where ${\bm X}_{i}^{j}\sim{\mathcal P}^{j}$ for $i=1,\ldots, N_{\!s}^{j}$ and $j=1,\ldots, N_{\!p}$.
Here, $N = \sum_{j=1}^{N_{\!p}} {N_{\!s}^{j}}$ is the total number of sampled paths, and the reweighting function $\gamma^{j}:\Omega \rightarrow{\mathbb R}$ can be any function that satisfies the relation
\begin{equation}
\ell({\bm X}) \neq 0 \quad \Rightarrow \quad \frac{1}{N}\! \sum_{j=1}^{N_{\!p}} {N_{\!s}^{j}} \gamma^{j}\! ({\bm X}) =1
\end{equation}
which guarantees that the resulting MC estimation $\hat{\rho}$ is unbiased~\cite{rubinstein2016simulation,thijssen2018consistent}. For example, the flat function $\gamma^{j}({\bm X}) =1$ for all ${\bm X}$ or the balance-heuristic function
$\gamma^{j}({\bm X}) = N / \sum_{k=1}^{N_{\!p}} \! N_{s}^{j} \frac{d{\mathcal P}^{k}}{d{\mathcal P}^{j}}({\bm X})$
can be employed~\cite{thijssen2018consistent}.

\subsection{Cross entropy method for PI: KL control}
\label{sec:disc:cross-entropy}
The well-known cross-entropy (CE) method~\cite{rubinstein2016simulation,rubinstein2004cross}, which was originally invented for derivative-free optimization, can also be applied to trajectory generation by computing the following information theory optimization:
\begin{equation}\label{eq:ce:1}
\begin{split}
\pi^* 
&= \arg \min_{\pi} D_{\rm KL}({\mathcal P}^*\|{\mathcal P}^{\pi}) \\
&= \arg \min_{\pi} {\mathbb E}_{{\mathcal P}^*} \!\! \left[ \ln \frac{d{\mathcal P}^*}{d{\mathcal P}^{\pi}} \right] \\
&= \arg \min_{\pi} {\mathbb E}_{{\mathcal P}^{\pi}} \!\! \left[ \frac{d{\mathcal P}^*}{d{\mathcal P}^{\pi}} \ln \frac{d{\mathcal P}^*}{d{\mathcal P}^{\pi}} \right] \\
&= \arg \min_{\pi} {\mathbb E}_{\!{\bm X}\sim{\mathcal P}^{\pi}} \!\! \left[ \omega^{\pi}\!({\bm X}) \ln \omega^{\pi}\!({\bm X}) \right] \\
&= \arg \min_{\pi} {\mathbb E}_{\!{\bm X}\sim{\mathcal P}^{\tilde \pi}} \!\! \left[ \omega^{\tilde\pi}\!({\bm X}) \ln \frac{\omega^{\pi}\!({\bm X})}{\omega^{\tilde\pi}\!({\bm X})} \right] \\
&= \arg \min_{\pi} {\mathbb E}_{\!{\bm X}\sim{\mathcal P}^{\tilde \pi}} \!\! \left[ {\omega}^{\tilde\pi}\!({\bm X})\ln {\omega}^{\pi}\!(\!{\bm X}\!) \right] 
\end{split}
\end{equation}
where the importance weight is defined as:
\begin{equation}\label{eq:importance_weight}
\omega^{\pi}\!({\bm X}) 
= \frac{d{\mathcal P}^*\!({\bm X})}{d{\mathcal P}^{\pi}\!({\bm X})} 
= \omega^{\tilde\pi}\!({\bm X}) \frac{d{\mathcal P}^{\tilde \pi}\!({\bm X})}{d{\mathcal P}^{\pi}\!({\bm X})}
\end{equation}
where $\tilde \pi$ is the baseline Markov policy.
In addition, rewriting the cost function in the fourth row of~\eqref{eq:ce:1} as $J^\pi({\bm X}) := {\omega}^{\pi}\!({\bm X})\ln {\omega}^{\pi}\!(\!{\bm X}\!)$,
we have the following expectation minimization: 
\begin{equation}\label{eq:ce:2}
\min_{\pi} {\mathbb E}_{\!{\bm X}\sim{\mathcal P}^{\pi}} \! \left[ J^\pi({\bm X}) \right] \,.
\end{equation}

Alg.~\ref{alg:ce} summarizes the iterative procedures of CE for motion planning~\cite{kobilarov2012cross} to solve the optimization problem ~\eqref{eq:ce:2} using a sampling-based method.

\begin{algorithm}[t]
	\caption{${\tt CE\_trajopt}$}\label{alg:ce}
	\begin{algorithmic}[1]
		\State \textbf{Input:} $K$: Number of samples
		\State $N$: Decision horizon
		\State $\pi^{0}$: Initial (trial) policy  
  
		\While{not converged}
  
		\State Sample trajectories $\{{\bm X}_{1}, \cdots, {\bm X}_{K}\}$ from ${\mathcal P}^{\pi^{i}}$.
		\State Determine the elite set threshold: $\gamma_{i} = J^{\pi^{i}}\!({\bm X}_{\kappa})$
		\State \quad where $\kappa$ denotes the index of the $K_{e}$ best 
		\State \quad sampled-trajectory with $K_{e} < K$.
		\State Compute the elite set of sampled-trajectories: 
		\State \quad $\displaystyle {\mathcal E}_{i}=\left\{{\bm X}_{k} | J^{\pi^{i}}\!({\bm X}_{k}) \leq \gamma_{i}\right\}$
		\State Update the policy:
		\State \quad
		$\pi^{i+1} = \displaystyle \arg \min_{\pi} \frac{1}{|{\mathcal E}_{i}|} \sum_{{\bm X}_{k} \in {\mathcal E}_{i}} J^{\pi}\!({\bm X}_{k})$
		\State Check convergence
		\EndWhile
	\end{algorithmic}
\end{algorithm}

\begin{remark}
{\it
For expectation minimization in~\eqref{eq:ce:2} and Alg.~\ref{alg:ce}, it is common to use a parameterization of the control policy $\pi$ or the resulting trajectory distribution ${\mathcal P}^{\pi}$ which can be rewritten as ${\mathcal P}^{\pi}\!({\bm X}) = {\mathcal P}({\bm X};\theta)$. This parameterization results in a finite-dimensional optimization. \hfill$\square$
}
\end{remark}

\subsection{MPC-like open-loop controller: MPPI}
\label{sec:PI:alg:mppi}
By applying time discretization with arithmetic manipulations to~\eqref{eq:picontrol}, the path integral control becomes
\begin{align*}
	u^*(t,x) = u(t,x) + \frac{{\mathbb E}_{\mathcal Q} [ \exp\left( -\frac{1}{\lambda}G_{t}^\pi \right)\! \delta u]}{{\mathbb E}_{\mathcal Q}[\exp\left( -\frac{1}{\lambda}G_{t}^\pi \right)]}
\end{align*}
where $u$ is the nominal control input and $\delta u$ is the deviation control input for exploration. Here, the expectation is considered with respect to the probability measure ${\mathcal Q}$ of a path corresponding to policy $\pi$.

For implementation, the expectation is approximated using MC importance sampling as follows:
\begin{align*}
	u^*(t,x) \approx u(t,x) + \sum_{k=1}^{K}  \frac{ \exp\!\left( -\frac{1}{\lambda}G_{t}^{\pi_k} \right)}
	{\sum_{\kappa=1}^{K} \exp\!\left( -\frac{1}{\lambda}G_{t}^{\pi_\kappa} \right)} \delta u_k (t,x)
\end{align*}
where $K$ is the number of sampled paths, and $G_{t}^{\pi_k}$ is the cost-to-go corresponding to the simulated trajectory following policy $\pi_{k}(t,x)=u(t,x)+\delta u_k (t,x)$ corresponding to the perturbed control inputs for $k=1,2,\ldots, K$. This path-integral control based on forward simulations is known as the model-predictive path integral (MPPI)~\cite{williams2017model,williams2019mppi,williams2016aggressive}. By recursively applying MPPI, the control inputs can approach the optimal points. Alg.~\ref{alg:mppi} summarizes the standard procedures for the MPPI.

\begin{algorithm}[t]
	\caption{${\tt MPPI\_control}$}\label{alg:mppi}
	\begin{algorithmic}[1]
		\State \textbf{Input:} $K$: Number of samples
		\State $N$: Decision horizon
		\State $(u_0,u_1,\dots,u_{N-1})$: Initial control sequence
		
		\While{not terminated}
		
		\State Generate random control variations $\delta u$
		
		\For{$k=1,\dots,K$}
		\State $x_0 = x_{\rm init}$
		\State $t_0 = t_{\rm init}$
		
		\For{$i=0,\dots,N-1$}
		\State $f_i = f(t_i,x_i)$
		\State $g_i = g(t_i,x_i)$
		\State $\tilde{u}_{i,k} = u_i + \delta u_{i,k}$
		\State $x_{i+1} = x_i + (f_i + g_i\tilde{u}_{i,k})\Delta t$
		\State $t_{i+1} = t_i + \Delta t$
		\EndFor
		
		\State $G_{N,k} = {\tt cost}(x_N)$
		\For{$i=N-1,\dots,0$}
		\State $G_{i,k} = G_{i+1,k} + {\tt cost}(x_i,\tilde{u}_{i,k})$
		\EndFor
		\EndFor
		
		\For{$i=0,\dots,N-1$}
		\State $w_{i,k} = \frac{\exp(-G_{i,k}/\lambda)}{\sum_{\kappa=1}^{K} \exp(-G_{i,\kappa}/\lambda)}$
		\State $u_i \leftarrow u_i +
		\sum_{k=1}^{K}w_{i,k}\delta u_{i,k}$
		\EndFor
		
		\State Send $u_0$ to actuators 
		\For{$i=0,\dots,N-1$}
		\State $u_i = u_{\min(i+1,N-1)}$
		\EndFor
		\State Update $x_{\rm init},t_{\rm init}$ by measurement
		\EndWhile
	\end{algorithmic}
\end{algorithm}

\subsection{Parameterized state feedback controller}
\label{sec:PI:alg:feedback}
Although MPC-like open-loop control methods are easy to implement, they exhibit certain limitations.
First, it can be inefficient for high-dimensional control spaces because a longer horizon results in a higher dimension of the decision variables. 
Second, it does not design a control law (i.e., policy), but computes a sequence of control inputs over a finite horizon, which means that whenever a new state is encountered, the entire computation should be repeated. Although a warm start can help solve this problem, it remains limited. 
Third, the trade-off between exploitation and exploration is not trivial.  

As an alternative to open-loop controller design, a parameterized policy or control law can be iteratively updated or learned via model-based rollouts, from which the performance of a candidate policy of parameterization is evaluated, and the parameters are updated to improve the control performance. The main computation procedure is that from an estimate of the probability $P(\bm{X}|x_0)$ of the sampled trajectories, we want to determine a parameterized policy $\pi_{t}(u_t | x_t; \theta_t)$ for each time $t < T$ that can reproduce the trajectory distribution $P(\bm{X}|x_0)$, where $\theta_t \in \Theta$ denotes the parameter vector that defines a feedback policy $\pi_{t}$~\cite{gomez2014policy,thijssen2015path}.
In general, a feedback policy can be time varying, and if it is time invariant, then the time dependence can be removed; that is, $\theta = \theta_t$ for all times $t \in [0,T]$. In this review paper, we consider only deterministic feedback policies, but the main idea can be trivially extended to probabilistic feedback policies\footnote{An example of probabilistic feedback policy parameterization is a time-dependent Gaussian policy that is linear in the states, $\pi_{t}(u_t | x_t; \theta_t) \sim \mathcal{N}(u_t | k_t + K_t x_t , S_t)$, in which the parameter vector is $\theta_t = (k_t, K_t, S_t)$ and updated by a weighted linear regression and the weighted sample-covariance matrix~\cite{kupcsik2013data,gomez2014policy}.}. 

\noindent
\textbf{Linearly parameterized state feedback} 
Consider 
\begin{equation}\label{eq:policy:para:1}
u(t,x) = h(t,x)^\top \theta
\end{equation}
where $h: {\mathbb R} \times {\mathbb R}^{n} \rightarrow {\mathbb R}^{n_{p}}$ is a user-defined feature of the state feedback control law. 
Using the model-based forward simulations, a control parameter update rule can be applied as follows:
\begin{equation}
\theta \leftarrow \theta + \sum_{k=1}^{K} w_{k} \delta \theta_{k}
\end{equation}
where $w_{k} = \frac{\exp\left( -\frac{1}{\lambda}G_{t}^{\pi_k} \right)}{\sum_{\kappa=1}^{K}\exp\left( -\frac{1}{\lambda}G_{t}^{\pi_\kappa} \right)}$ is the weight for the $k$th sampled perturbation-parameter $\delta \theta_{k}$ of the linearly parameterized control law and ${\pi_k} (t,x)= h(x,t)^\top \!(\theta+ \delta \theta_{k})$ is the test or the exploration (search) policy.

\noindent
\textbf{Nonlinearly parameterized state feedback} 
Consider 
\begin{equation}\label{eq:policy:para:2}
u(t,x) = \pi(t,x;\theta) 
\end{equation}
where the state-feedback control law is parameterized by the control parameter $\theta \in {\mathbb R}^{n_{p}}$.
Using the model-based forward simulations, a control parameter update rule can be applied as follows:
\begin{equation}
\theta \leftarrow \theta + \sum_{k=1}^{K} {\tilde w}_{k} \delta \theta_{k}
\end{equation}
where the weight is defined as follows:
\begin{equation}
{\tilde w}_{k} = w_{k} [\nabla_{\theta} \pi(t,x;\theta)]^\dagger [\nabla_{\theta} \pi(t,x;\theta+\delta\theta_{k})]
\end{equation}
with the weight $w_{k} = \frac{\exp\left( -\frac{1}{\lambda}G_{t}^{\pi_k} \right)}{\sum_{\kappa=1}^{K}\exp\left( -\frac{1}{\lambda}G_{t}^{\pi_\kappa} \right)}$ and the exploration policy $\pi_{k}= \pi(t,x;\theta+\delta\theta_{k})$. Here, $[\cdot]^\dagger$ denotes the pseudoinverse. 

\begin{remark}[CE method for policy improvement]
{\it
Parameterized state feedback controls, such as~\eqref{eq:policy:para:1} and \eqref{eq:policy:para:2}, can also be updated using CE methods.
For example, $\delta \theta_{k} \sim {\mathcal GP}(0,\Sigma)$ is samples, the cost of simulated trajectories is evaluated with control parameters~$\theta_{k} = \theta + \delta \theta_{k}$~for~$k=1,\ldots,K$, the samples are sorted in ascending order according to the simulated costs, and the parameter is updated by weighted-averaging the sampled parameters $\delta \theta_{k}$ from the sorted elite set, $\theta \leftarrow \theta + {\tt average}(\delta\theta_{k})_{\tt elite}$. In general, the covariance $\Sigma$ can be also updated by empirical covariance of the sampled parameters $\delta \theta_{k}$ from the sorted elite set, $\Sigma \leftarrow \Sigma + {\tt average}(\delta\theta_{k}\delta\theta_{k}^\top)_{\tt elite}$. \hfill$\square$
}
\end{remark}

\subsection{Policy improvement with path integrals}
\label{sec:PI:alg:PI2}
%
%
%
%
%
%

Policy improvement with path integrals (PI$^{2}$) is presented in~\cite{theodorou2010generalized}. The main idea of PI$^{2}$ is to iteratively update the policy parameters by averaging the sampled parameters in weights with the costs of the path integral corresponding to the simulated trajectories~\cite{yamamoto2020path}. 
Alg.~\ref{alg:PI2CMA}~shows the pseudocode for the PI$^{2}$ Covariance Matrix Adaptation (PI$^{2}$-CMA) proposed in~\cite{stulp2012path} based on the CMA evolutionary strategy (CMAES)~\cite{hansen2001completely,hansen2016cma}. In~\cite{stulp2012path}, the PI$^{2}$-CMA was compared with CE methods and CMAES in terms of optimality, exploration capability, and convergence rate. 
Skipping the covariance adaptation step in Alg.~\ref{alg:PI2CMA} yields a vanilla PI$^{2}$. 
In~\cite{theodorou2010generalized}, it was also shown that policy improvement methods based on PI$^{2}$ would outperform existing gradient-based policy search methods such as REINFORCE and NAC.

\begin{algorithm}[t]
	\caption{${\tt PI^{2}\mbox{-}CMA}$}\label{alg:PI2CMA}
	\begin{algorithmic}[1]
		\State \textbf{Input:} $K$: Number of samples
		\State $N$: Decision horizon
		\State $(\theta,\Sigma)$: Initial hyper-parameter
		
		\While{not terminated}
		
		\State Generate random variables $\theta_{i,k}\sim{\mathcal GP}(\theta,\Sigma)$
		\State Generate random initial conditions $x_{\rm init}$
		
		\For{$k=1,\dots,K$}
			\State $x_0 = x_{\rm init}$
			\State $t_0 = t_{\rm init}$
		
			\For{$i=0,\dots,N-1$}
				\State $f_i = f(t_i,x_i)$
				\State $g_i = g(t_i,x_i)$
				\State ${u}_{i,k} = \pi(t_i, x_i; \theta_{i,k})$
				\State $x_{i+1} = x_i + (f_i + g_i {u}_{i,k})\Delta t$
				\State $t_{i+1} = t_i + \Delta t$
			\EndFor
		
			\State $G_{N,k} = {\tt cost}(x_N)$
			
			\For{$i=N-1,\dots,0$}
				\State $G_{i,k} = G_{i+1,k} + {\tt cost}(x_i,{u}_{i,k})$
			\EndFor
			
		\EndFor
		
		\For{$i=0,\dots,N-1$}\vspace{1mm}
			\State $w_{i,k} = \frac{\exp(-G_{i,k}/\lambda)}{\sum_{\kappa=1}^{K} \exp(-G_{i,\kappa}/\lambda)}$\vspace{1mm}
			\State $\theta_i = \sum_{k=1}^{K} w_{i,k} \theta_{i,k}$\vspace{1mm}
			\State $\Sigma_{i} = \sum_{k=1}^{K} w_{i,k} (\theta_{i,k}\!-\theta) (\theta_{i,k}\!-\theta)^\top$ \vspace{1mm}
		\EndFor \vspace{1mm}
		
		\State $\theta \leftarrow \sum_{i=0}^{N-1}\frac{N-i}{\sum_{j=0}^{N-1}(N-j)} \theta_{i}$ \vspace{1mm}
		\State $\Sigma \leftarrow \sum_{i=0}^{N-1}\frac{N-i}{\sum_{j=0}^{N-1}(N-j)} \Sigma_{i}$
		
		\EndWhile
	\end{algorithmic}
\end{algorithm}

\section{Python and MATLAB Simulation Results}
\label{sec:sim:matlab}
This section presents the simulation results of two different trajectory optimization problems: 1D cart-pole trajectory optimization and bicycle-like mobile robot trajectory tracking. The first case study demonstrates the simulation results for the cart-pole system, followed by the second numerical experiment showing the local trajectory tracking of a bicycle-like mobile robot with collision avoidance of dynamic obstacles\footnote{Videos of simulation results are available at
\begin{itemize}
\item
\href{https://youtu.be/zxUN0y23qio}{https://youtu.be/zxUN0y23qio}
\item
\href{https://youtu.be/LrYvrgju_o8}{https://youtu.be/LrYvrgju\_o8}
\end{itemize}
}.
The details of the simulation setups and numerical optimal control problems are not presented here because of limited space, but they are available in the accompanying github page\footnote{\href{https://github.com/INHA-Autonomous-Systems-Laboratory-ASL/MATLAB-Simulation-An-Overview-of-Recent-Advances-in-Path-Integral-Control.git}{https://github.com/iASL/pic-review.git}}.

\subsection{Cart-pole trajectory optimization}
\label{sec:numexp:2:cart-pole}
In this section, we consider trajectory optimization for set-point tracking in a one-dimensional (1D) cart-pole system. This system comprises an inverted pendulum mounted on a cart capable of moving along a unidirectional horizontal track. The primary goal is to achieve a swing-up motion of the pole and subsequently maintain its stability in an upright position through horizontal movements of the cart. We considered a quadratic form for the cost function associated with each state variable. To address this challenge of stabilizing a cart-pole system, we implemented several path integral methods including CEM, MPPI, and  PI$^{2}$-CMA delineated in Algs.~\ref{alg:ce},~\ref{alg:mppi}, and \ref{alg:PI2CMA}, and investigated their performances with comparison to nonlinear model predictive control (NMPC). 

\begin{figure}[t]
	\centering
	\includegraphics[width=1.0\linewidth]{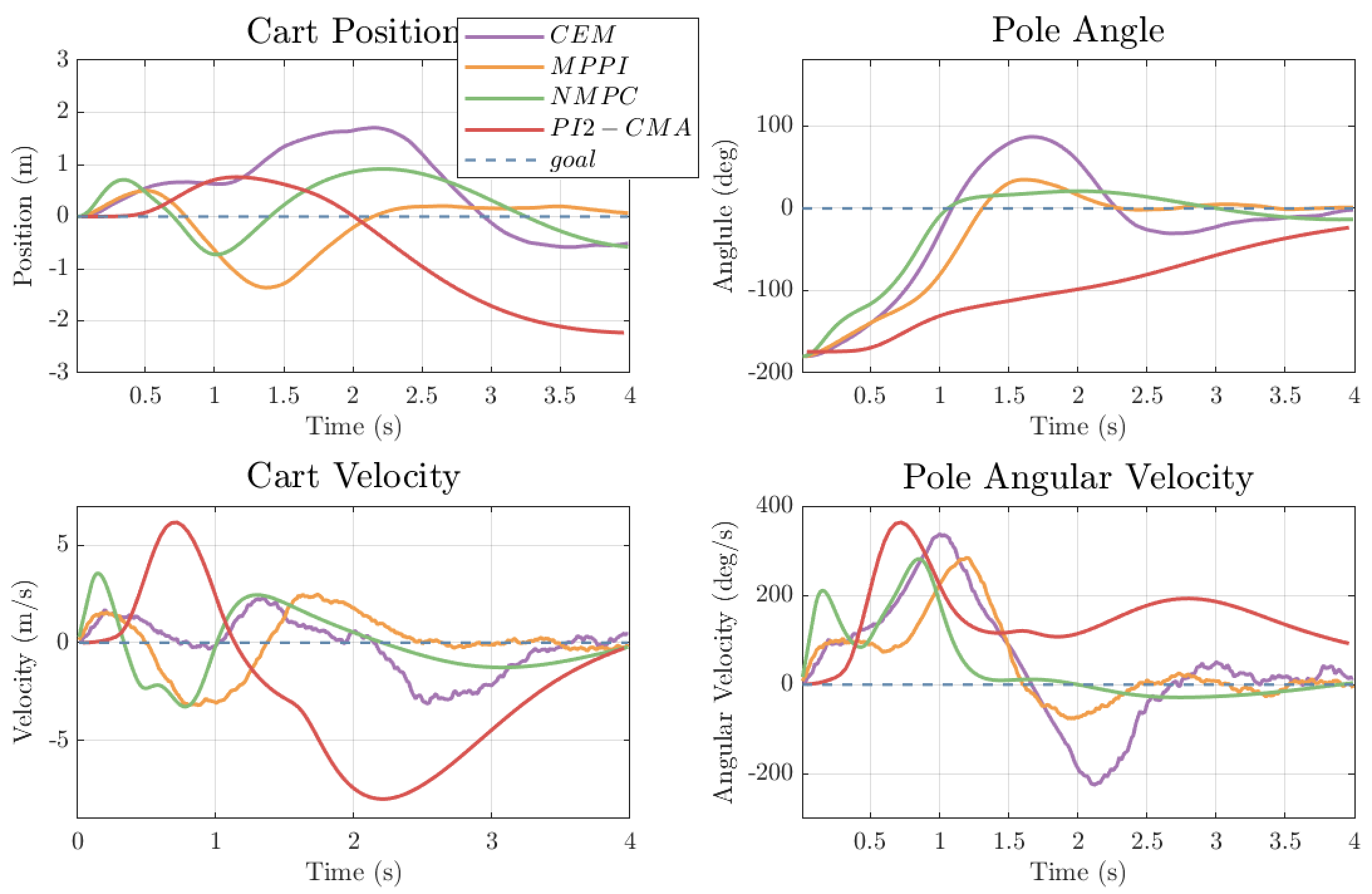}\vspace{-3mm}
	\caption{Controlled trajectories of the 1D cart-pole using different methods of path integral control (CEM, MPPI,PI$^2$-CMA) in comparison with nonlinear model predictive control (NMPC).}
	\label{fig:cart_pole_pose}
\vspace{3mm}
	\centering
	\includegraphics[width=1.0\linewidth]{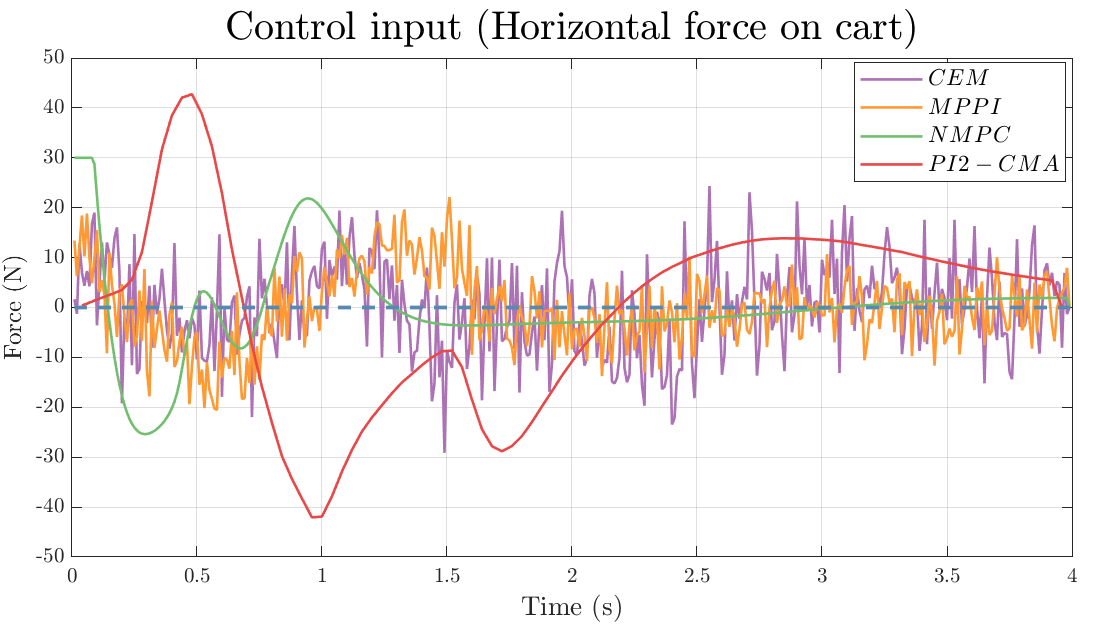}\vspace{-3mm}
	\caption{Control inputs resulting from different methods of path integral control (CEM, MPPI,PI$^2$-CMA) and NMPC.}
	\label{fig:control_cart_pole}
\vspace{3mm}
	\centering
	\includegraphics[width=1.0\linewidth]{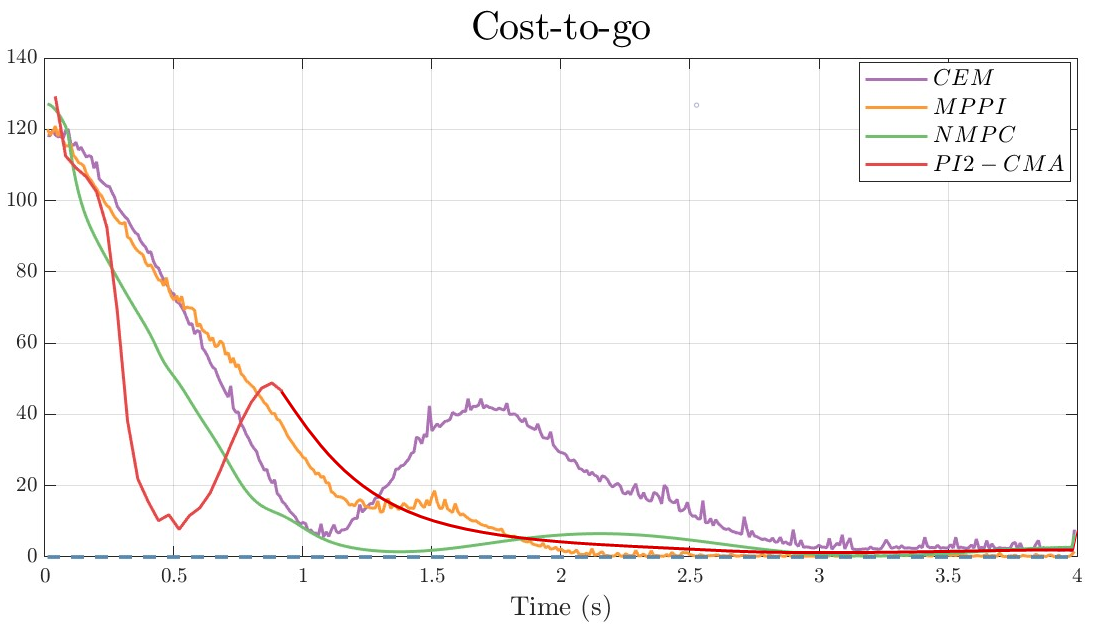}\vspace{-3mm}
	\caption{Cost-to-go associated with different methods of path integral control (CEM, MPPI,PI$^2$-CMA) and NMPC.}
	\label{fig:cost_to_go_cart_pole}
\end{figure}

Simulations are performed in Python environments. 
Fig.~\ref{fig:cart_pole_pose} illustrates the controlled trajectories of the cart-pole system under various control strategies. Fig.~\ref{fig:control_cart_pole} depicts the force inputs derived from the application of each control method. These results demonstrate the effectiveness of the path integral control methods in achieving the desired control objectives for the cart-pole system.
The comparative analysis of our results reveals distinct characteristics and performance efficiencies among the different path integral control methods. MPPI and CEM methods achieved better performance of stabilizing the pole as compared to NMPC and PI$^{2}$-CMA. Upon fine-tuning optimization parameters, MPPI achieved faster convergence in pole stabilization than CEM, given the same prediction horizon and sample size. Such enhanced performance of MPPI is further corroborated by its superior cost-to-go metrics, as shown in Fig.~\ref{fig:cost_to_go_cart_pole}.

This comparison of four different PI-based control strategies stabilizing a cart-pole system reveals notable characteristics. CEM and MPPI result in smooth trajectories of the cart-pole position and lower accumulated cost-to-go, indicating better energy efficiency. In contrast, PI$^{2}$-CMA and NMPC result in more aggressive control input sequences with faster responses and larger overshoots in velocities, which might increase energy use and hasten actuator degradation. However, CEM and MPPI show higher rate of input changes, which would not be desirable for real hardware implementations. Of course, this jerking behavior of control actions can be relaxed by putting rate or ramp constraints in control inputs. The appropriate choice among these PI-based control strategies would heavily depend on the specific demands of the application, balancing multiple objectives such as efficiency, responsiveness, and operational durability. 

\subsection{Path planning for bicycle-like mobile robot}
\label{sec:numexp:3:quadrotor2}
The robot system we consider in this section is a mobile robot navigating in an environment with obstacles, aiming to follow a predefined path while avoiding collisions and staying inside a track. To evaluate the effectiveness of an MPPI controller for real-time trajectory generation with obstacle avoidance, we conducted a series of MATLAB simulations where two moving obstacles are considered in the 2D coordinates.

\begin{figure}[t]
	\centering
	\includegraphics[width=.875\linewidth]{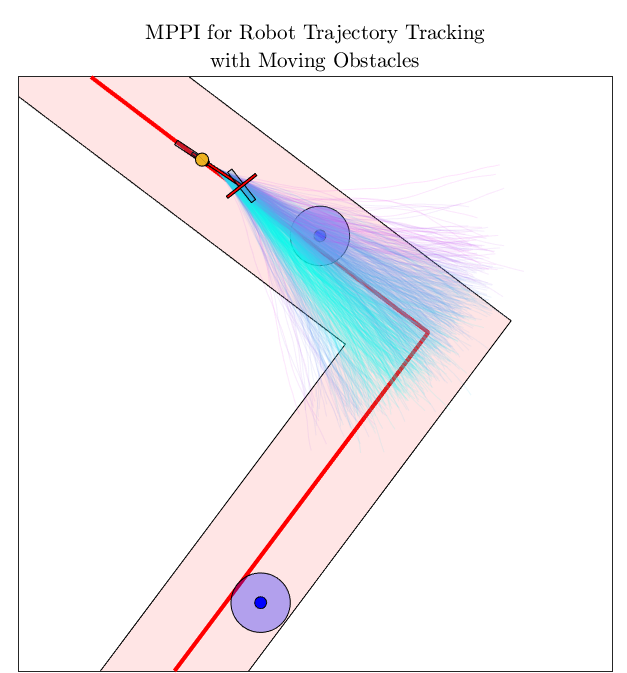}\vspace{-2mm}
	\caption{A capture of simulating local path planning and tracking with obstacle avoidance using an MPPI controller. An associated video of simulations is available at~\href{https://youtu.be/LrYvrgju_o8}{https://youtu.be/LrYvrgju\_o8}.}
	\label{fig:Robot_MPPI}
\vspace{3mm}
	\centering
	\includegraphics[width=1.0\linewidth]{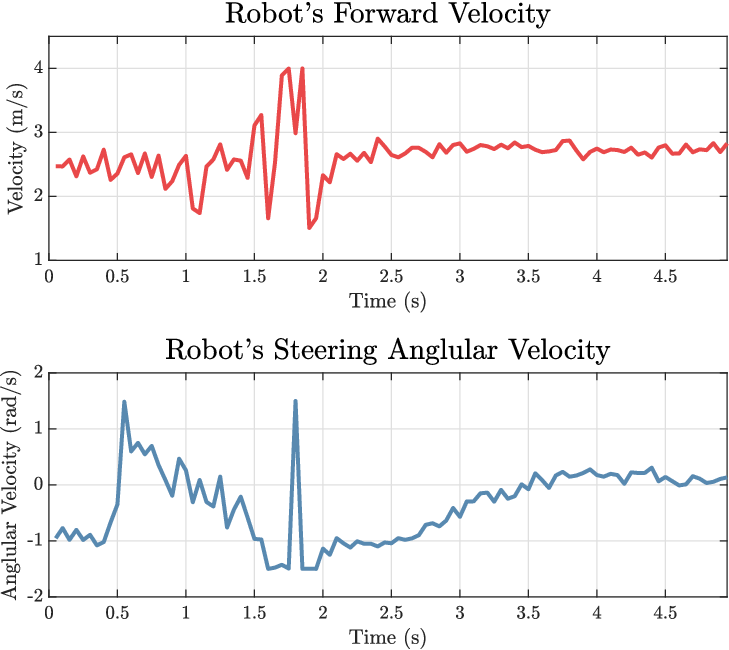}\vspace{-3mm}
	\caption{Forward and angular velocity trajectories resulting from a MPPI-based controller steering a bicycle-like mobile robot.}
	\label{fig:Robots_angle_velocity}
\end{figure}

We implemented the MPPI controller described in Alg.~\ref{alg:mppi} for bicycle-like mobile robot navigation over a track. Fig.~\ref{fig:Robot_MPPI} shows the robot's trajectory tracking while avoiding moving obstacles. Fig.~\ref{fig:Robots_angle_velocity} shows the forward and angular velocities of the robot required to follow the desired path by avoiding obstacles, which assesses the effectiveness of the controller in avoiding obstacles.
The results demonstrated that the MPPI controller achieved successful trajectory generation and tracking, while effectively avoiding dynamic obstacles. Throughout the simulation of mobile robot path planning, MPPI controller showed robust obstacle avoidance capabilities, successfully navigating around obstacles, and minimizing the distance with its reference.

\section{MPPI-based Autonomous Mobile Robot Navigation in ROS2/Gazebo Environments}
\label{sec:sim:gazebo}
In this section, we consider two ROS2/Gazebo simulations of MPPI-based autonomous mobile robot navigation in a cafeteria environment and in a maze. We implement various path integral algorithms including MPPI, Smooth MPPI, and Log MPPI to enhance navigation and control in these complex scenarios. The SLAM toolbox is employed for mapping while Navigation2 (NAV2) is used for navigation within the ROS2 Gazebo simulation environment~\cite{macenski2022robot,macenski2023desks}. These path integral control approaches are attractive because they are derivative-free and can be parallelized efficiently on both CPU and GPU machines. The simulations are conducted using a computing system equipped with Intel Core i7 CPU and NVIDIA GeForce RTX 3070 GPU. These simulations are executed within the Ubuntu 22.04 operating system, leveraging the ROS2 Humble simulation platform for comprehensive analysis. The following subsections provide the details of the simulation setups and results of different MPPI-based path planning algorithms in two different scenarios. The simulation frameworks and source codes used in this study are publicly available\footnote{\href{https://github.com/INHA-Autonomous-Systems-Laboratory-ASL/MATLAB-Simulation-An-Overview-of-Recent-Advances-in-Path-Integral-Control.git}{https://github.com/iASL/pic-review.git}}. 

\subsection{Autonomous robot navigation in a cafeteria environment}
In this subsection, we present the simulation results of the autonomous robot navigation using different MPPI-based path planning algorithms in a cafeteria environment. The primary objective of this ROS2/Gazebo simulation is to evaluate the performance of MPPI-based path planning algorithms for a task of navigating an indoor autonomous mobile robot serving foods at different tables while avoiding obstacles in a confined and dense indoor space.

\subsubsection{Experiment setup}
\begin{figure}[!t]
	\centering
	\includegraphics[width=\linewidth]{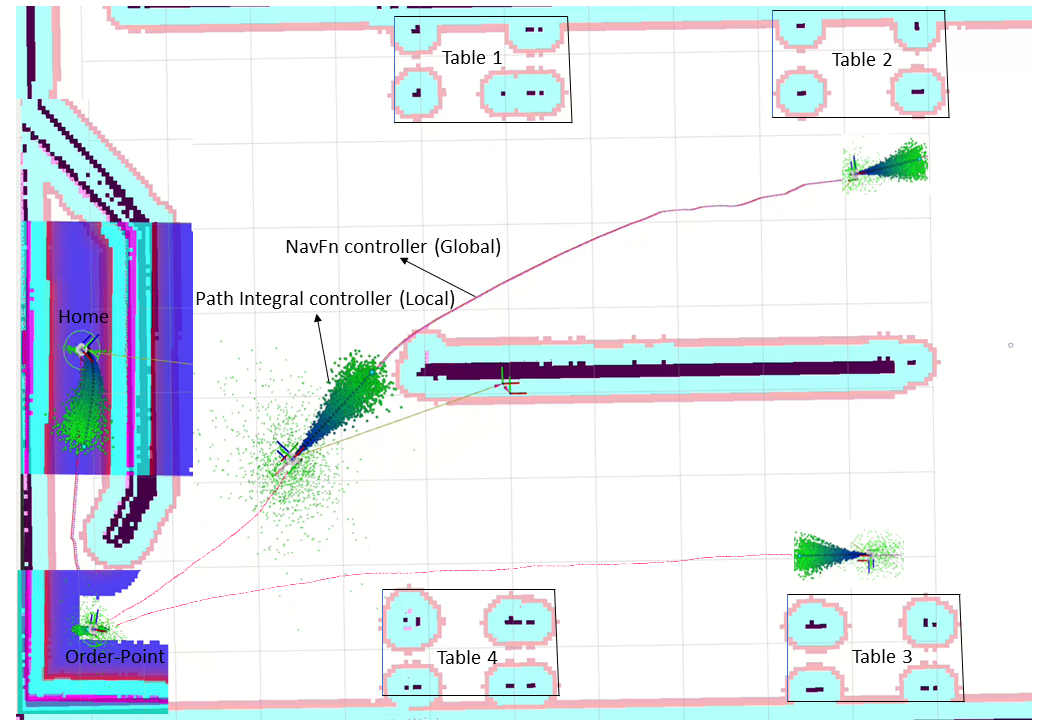}\vspace{-2mm}
	\caption{ROS2/Gazebo simulation navigating an autonomous mobile robot in a cafeteria environment for which MPPI, Smooth MPPI, and Log MPPI are applied as local controllers.}
	\label{fig:Indoor_cafe}
\vspace{4mm}
	\centering
	\includegraphics[width=\linewidth]{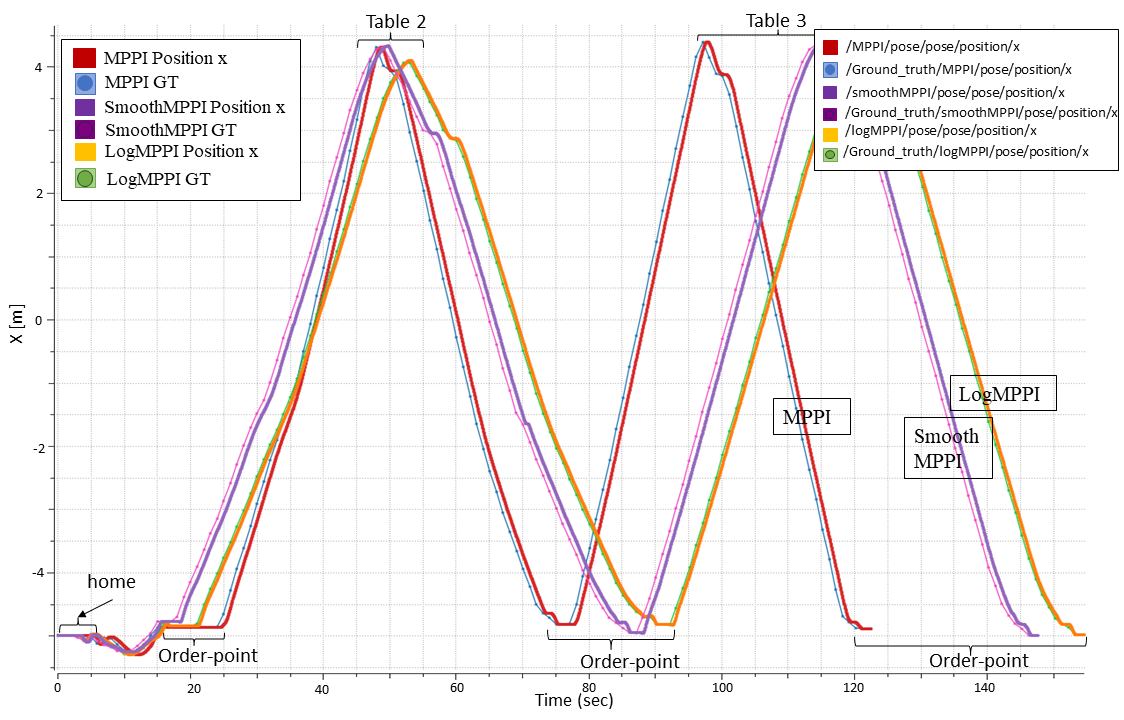}\vspace{-3mm}
	\caption{Performance comparison of path integral controllers in terms of the longitudinal position $(x)$.}
	\label{fig:pathintegral_hotel_position}
\vspace{4mm}
	\centering
	\includegraphics[width=\linewidth]{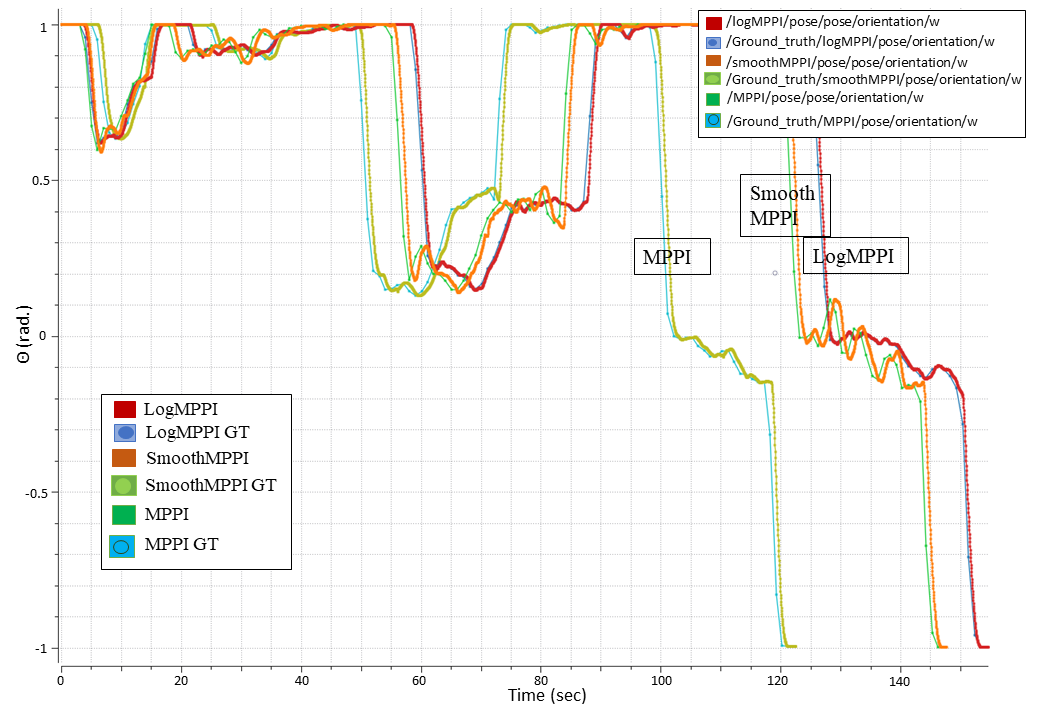}\vspace{-3mm}
	\caption{Performance comparison of path integral controllers in terms of the orientation $(\theta)$.}
	\label{fig:pathintegral_hotel_orientation}
\end{figure}

For the autonomous indoor robot navigation scenario, we utilize an autonomous mobile robot equipped with the necessary sensors such as a LiDAR and a camera for environmental perception. The cafeteria environment is designed using a Gazebo, accurately replicating the challenges of indoor navigation, including cluttered spaces, narrow passages, and obstacles along a path generated by a global planner, such as {\tt NavFn}, as shown in Fig.~\ref{fig:Indoor_cafe}. All three path integral controllers run in a receding-horizon fashion with $100$-time steps of prediction horizon where each time-step corresponds to $0.1$ sec. The number of control rollouts is $1024$ and the number of sampled traction maps is $2000$ at the rate of $30$ Hz. It can also replan at $30$ Hz while sampling new control actions and maps. 

In the simulated cafeteria scenario demonstrating the navigational efficiency in a dynamic service environment, as shown in Fig.~\ref{fig:Indoor_cafe}, the robot is tasked to navigate a series of waypoints representing a service cycle. Commencing from the home coordinates $(x = -5, y = 0.51, \theta = 0.01)$, the robot is required to proceed to the order-taking waypoint $(x = -4.85, y = -3.0, \theta = 0.01)$. Subsequently, it traverses to serve table 2 at the coordinates $(x = 4.29, y = 2.64, \theta= 0.01)$ and upon order collection, returned to the order-taking waypoint. The mission is completed when the robot serves table 3 at the coordinates $(x = -0.6, y = -1.99, \theta = 0.01)$ and returns to the order-taking location. The total navigational distance required for the service unit to traverse, beginning from its home location, involves a multi-point itinerary. Initially, it must travel to the order point, followed by a journey to table 2. Afterwards, it returns to the order point, from where it proceeds to serve table 3, and subsequently returns to the order-taking point. This entire route encompasses a distance of approximately 43.15 meters.

\subsubsection{Simulation results}
The simulation results of autonomous robot navigation in a cafeteria environment are given in Fig.~\ref{fig:Indoor_cafe} and the accompanying video\footnote{\href{https://youtu.be/3VChYScJ7oA}{https://youtu.be/3VChYScJ7oA}}, to illustrate and compare the effectiveness of the three path integral controllers, MPPI, Smooth MPPI, and Log MPPI. 
In the simulated hotel navigation task, the performance of an autonomous delivery robot is evaluated using the {\tt NavFn} global planner, in conjunction with three local planners: MPPI, Smooth MPPI, and Log MPPI. Figs.~\ref{fig:pathintegral_hotel_position} and~\ref{fig:pathintegral_hotel_orientation} demonstrate the longitudinal position and orientation tracking of the robot pose over time.

In the Gazebo simulations captured in Fig.~\ref{fig:Indoor_cafe}, all three path integral controllers demonstrate remarkable indoor navigation capabilities for the differential-driving mobile robot, Turtlebot3. All three MPPI-based mobile robot navigation result in in smooth trajectory tracking while effectively avoiding obstacles and collisions in crowded café environments. 
Fig.~\ref{fig:pathintegral_hotel_position} shows the performance of the robot's positional components plotted over time.  Each trajectory—MPPI (red line), Smooth MPPI (dark blue line), and Log MPPI (yellow line)—is compared against the ground truth (dotted\_light red, dotted dark blue, and dotted green lines respectively). Despite slight deviations, the path integral controllers maintained a closely matched orientation with the ground truth, signifying robust direction control. Fig.~\ref{fig:pathintegral_hotel_orientation} shows the robot's orientation $(\theta)$ as guided by the {\tt NavFn} global planner with the three local planners. Similar to Fig.~\ref{fig:pathintegral_hotel_position}, the tracking performance of the three algorithms is compared against the ground truth. In Figs.~\ref{fig:pathintegral_hotel_position} and~\ref{fig:pathintegral_hotel_orientation}, the legends positioned in the top right corner display the Robot Operating System (ROS) topics pertaining to the robot's odometry and Gazebo ground truth data.

In the simulation study shown in Figs.~\ref{fig:Indoor_cafe},~\ref{fig:pathintegral_hotel_position},~and \ref{fig:pathintegral_hotel_orientation}, all MPPI-based path integral algorithms demonstrated high precision in trajectory tracking. Notably, Smooth MPPI (represented by a dark blue line) maintained a trajectory closest to the ground truth, especially in the simulation's latter stages. While MPPI excelled in speed, taking less time, it exhibited higher positional and angular errors compared to Smooth MPPI and Log MPPI. Log MPPI required marginally more time than Smooth MPPI. However, in terms of both positional accuracy and angular orientation, Smooth MPPI outperformed both Log MPPI and MPPI. This highlights Smooth MPPI's superior balance of speed and precision in trajectory tracking in a cafeteria environment.

\subsection{Autonomous robot navigation in maze-solving environment}
In this subsection, we detail the simulation outcomes of utilizing the different path integral controllers for the navigation of the robot in a maze-solving environment. The primary goal of a maze-solving robot is to autonomously navigate through a labyrinth from a starting point to a designated endpoint in the shortest time possible. It must efficiently map the maze, identify the optimal path, and adjust to obstacles using its onboard sensors and algorithms. 

\subsubsection{Experimental setup}
In our experiment, a simulated maze-solving robot was deployed in a ROS2/Gazebo environment, equipped with LiDAR and IMU sensors for navigation. The robot utilized the {\tt NavFn} global planner for overall pathfinding and local controllers including MPPI, Smooth MPPI, and Log MPPI for fine-tuned maneuvering. Performance was assessed based on the robot's ability to discover the most efficient path to the maze center, measuring metrics such as completion time and path optimality. The path integral controllers ran in a receding-horizon fashion with 100-time steps; each step was 0.1 s. The number of control rollouts was 1024, and the number of sampled traction maps was 2000 at a rate of 30 Hz. It could also replan at 30 Hz while sampling new control actions and maps. The computational overhead of these algorithms remained reasonable, thereby ensuring real-time feasibility for practical applications.

\begin{figure}[!t]
	\centering
	\includegraphics[width=\linewidth]{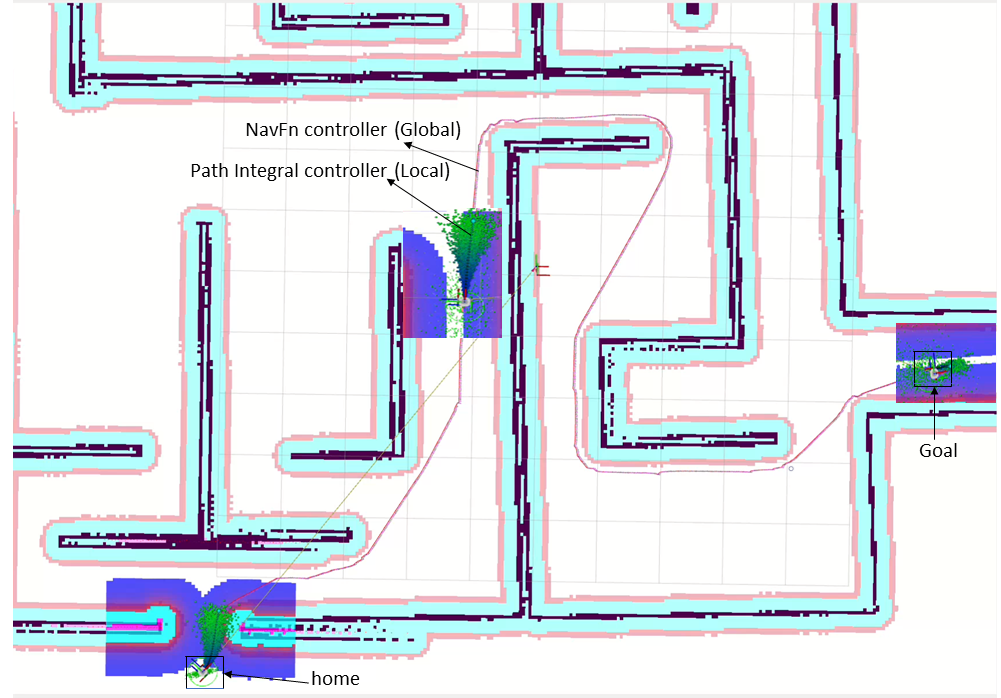}\vspace{-2mm}
	\caption{ROS2/Gazebo simulation navigating an autonomous mobile robot in a maze-solving environment for which MPPI, Smooth MPPI, and Log MPPI are applied as local controllers.}
	\label{fig:maze_env}
\vspace{4mm}
	\centering
	\includegraphics[width=\linewidth]{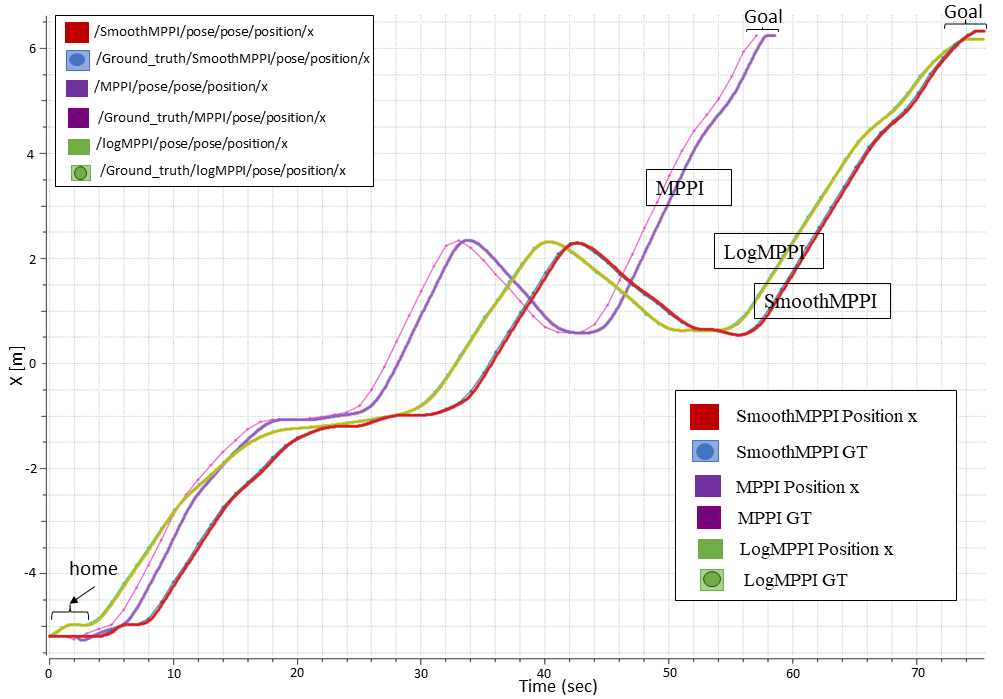}\vspace{-3mm}
	\caption{Performance comparison of path integral controllers in terms of the longitudinal position $(x)$.}
	\label{fig:maze_position}
\vspace{4mm}
	\centering
	\includegraphics[width=\linewidth]{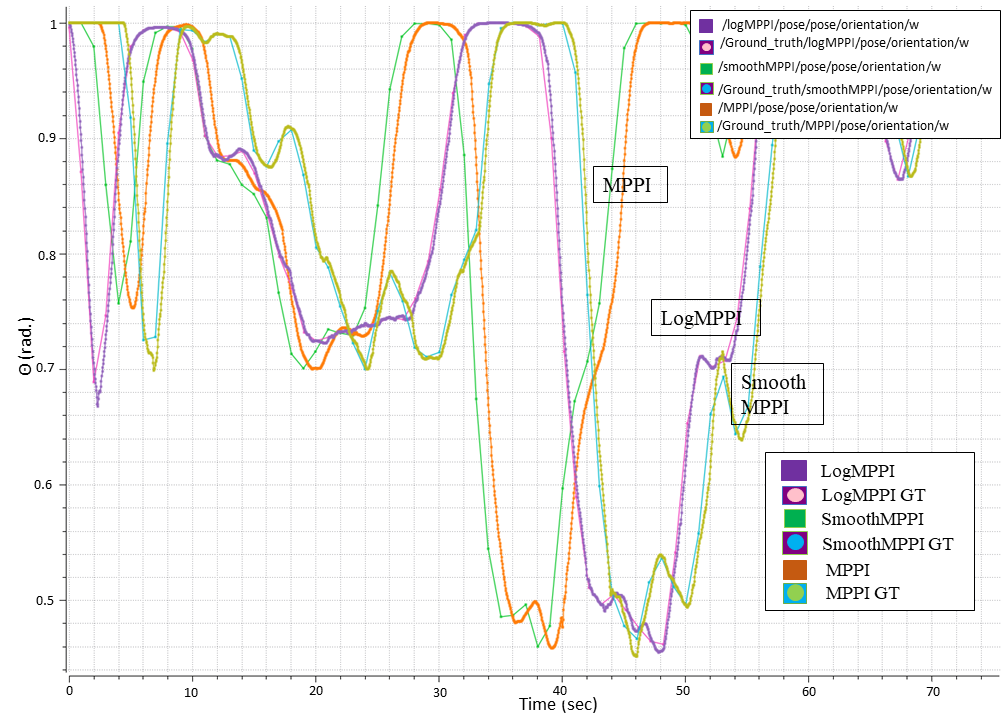}\vspace{-3mm}
	\caption{Performance comparison of path integral controllers in terms of the orientation $(\theta)$.}
	\label{fig:maze_orientation}
\end{figure}

In the simulated cafeteria scenario, as shown in Fig.~\ref{fig:maze_env}, the autonomous robot was programmed to navigate from an initial location at coordinates $(x = -5.18, y = -6.58, \theta = 0.99)$ to a predetermined destination $(x = 6.25, y = -1.47, \theta = 0.99)$. The cumulative distance from the starting home location to the designated end point is approximately 26.28 meters. The objective was to optimize the route for the shortest transit time, showcasing the robot's pathfinding proficiency in a complex simulated labyrinth.

\subsubsection{Simulation results}
For the maze-solving task, the robots were similarly directed by the {\tt NavFn} global planners, with the local planning algorithms tested for their path optimization efficiency. In the ROS2/Gazebo simulation, the maze-solving robot successfully navigated to the maze's center, with the {\tt NavFn} global planner and MPPI local controller achieving the most efficient path in terms of time and distance. The Smooth MPPI and Log MPPI controllers also completed the maze with competitive times, demonstrating effective adaptability and robustness in pathfinding within the complex simulated environment. Fig.~\ref{fig:maze_env} and a simulation video\footnote{\href{https://youtu.be/GyKDP3-NYA0}{https://youtu.be/GyKDP3-NYA0}} demonstrate the efficacy of the path integral controllers in navigating the maze-solving environment.

\begin{table}[t]
\centering
\begin{tabular}{|@{\,\,\,}c@{\,\,\,}||@{\,\,\,}c@{\,\,\,}|@{\,\,\,}c@{\,\,\,}|@{\,\,\,}c@{\,\,\,}|}
\hline
\multirow{2}{*}{Controller} & \multirow{2}{*}{Env.} & \multirow{2}{*}{Mission Time (s)}  & \multirow{2}{*}{Tracking Error ($\ell_{\infty}$)} \\
 & & & \\
\hline
\multirow{2}{*}{MPPI} & Cafe & 122  & 0.21 \\[.5mm]
 & Maze & 59 & 0.25 \\[1mm]
\hline 
\multirow{2}{*}{Smooth MPPI} & Cafe & 147  & 0.14 \\[.5mm]
 & Maze & 75 & 0.25 \\[1mm]
\hline
\multirow{2}{*}{Log MPPI} & Cafe & 155 & 0.1 \\[.5mm]
 & Maze & 75  & 0.25 \\[1mm]
\hline
\end{tabular}\vspace{2mm}
\caption{Comparative performance metrics of different path integral controllers across different environments}
\label{tab:controller_performance}
\end{table}

Fig.~\ref{fig:maze_orientation} illustrates the robot's orientation in a maze-solving environment. The ground truth data (dotted lines) represents the ideal orientation path for maze navigation. It is evident from the graph that the Smooth MPPI (green line) and Log MPPI (blue line) algorithms maintained a consistent orientation close to the ground truth, while the MPPI (red line) displayed marginally increased variance. The robot's $x$-position component within in maze-solving environment is shown in Fig.~\ref{fig:maze_position}, in which the ground truth path (dotted line) is tightly followed by all three algorithms. The Smooth MPPI (red line) and Log MPPI (green line) displayed almost identical performance, with MPPI (blue line) showing slight divergence yet still within an acceptable range for effective maze navigation. In evaluating speed and time efficiency, MPPI demonstrated superior performance, achieving the goal more swiftly compared to Log MPPI and Smooth MPPI. In Fig.~\ref{fig:maze_position}, the legend in the top left corner illustrates the ROS topics of the robot's odometry, whereas in Fig.~\ref{fig:maze_orientation}, the top right corner legend presents the ROS topics associated with Gazebo ground truth data.

The simulation results demonstrate that all considered path integral algorithms MPPI, Smooth MPPI, and Log MPPI perform robustly in trajectory optimization tasks for both hotel navigation and maze-solving scenarios. The angular orientation and position tracking graphs indicate that each algorithm is capable of closely following a predetermined ground truth trajectory with minimal deviation. However, minor differences in performance suggest that certain algorithms may be more suitable for specific applications, as given in Table~\ref{tab:controller_performance}. For instance, Log MPPI and Smooth MPPI's trajectory in the hotel navigation simulation suggests a potential for finer control in more predictable environments like cafeteria or warehouse environments, whereas MPPI showed promising results in the more dynamic and fast speed tracks or environments like maze-solving context or F1Tenth racing car competitions.

\section{Discussion and Future Directions}\label{sec:discussion}

\subsection{Policy parameterization}
\label{sec:disc:parapolicy}
There are multiple ways of policy parameterizations for state-feedback~\cite{deisenroth2013survey}. For example, 
linear policies~\cite{deisenroth2013survey,vinogradska2020quadrature}, radial basis function (RBF) networks~\cite{deisenroth2013gaussian,deisenroth2013survey,thor2020generic,vinogradska2020quadrature}, and dynamic movement primitives (DMPs)~\cite{schaal2005learning,ijspeert2013dynamical}~have been commonly used for policy representations and search in robotics and control. 

Note that because the PI-based control and planning algorithms presented in Section~\ref{sec:PI:alg} are derivative-free, complex policy parameterization and optimization can be implemented without any additional effort (e.g., numerical differentiation computing gradients, Jacobians, and Hessians). This is one of the advantageous characteristics of sampling-based policy search and improvement methods such as PI. 

\subsection{Path integral for guided policy search}
\label{sec:disc:policy-search}
The guided policy search (GPS), first proposed in~\cite{levine2013guided}, is a model-free policy-based reinforcement learning (RL). 
Pixel-to-torque end-to-end learning of visuomotor policies has recently become popular for RL in robotics~\cite{levine2016end}.
Compared with direct deep reinforcement learning, GPS has several advantages, such as faster convergence and better optimality.
In the GPS, learning consists of two phases. The first phase involves determining a local guiding policy, in which a training set of controlled trajectories is generated. In the second phase, a complex global policy is determined via supervised learning, in which the expected KL divergence of the global policy from the guiding policy is minimized. The goal of the GPS framework is to solve an information-theoretic-constrained optimization of the following~\cite{montgomery2016guided}:
\begin{equation}\label{eq:MDGPS}
\begin{split}
\min_{\theta,\beta} \ \,  & {\mathbb E}_{\beta}[G({\bm X})] \\
\mbox{s.t.}            \ \, & D_{\rm KL} (\beta({\bm X})\| \pi({\bm X};\theta)) \leq \epsilon
\end{split}
\end{equation}
where the KL divergence $D_{\rm KL} (\beta({\bm X})\| \pi({\bm X};\theta))$ can be rewritten as:
\[
\sum_{i=0}^{N-1} {\mathbb E}_{\beta}[ D_{\rm KL} (\beta(u_{i}|x_{i})\| \pi(u_{i}|x_{i};\theta)) ] \,.
\]
A baseline (Markovian) policy $\beta(u_{i}|x_{i})$ is a local guiding policy used to generate sampled trajectories starting with a variety of initial conditions.  
A parameterized policy $\pi(u_{i}|x_{i};\theta)$ is a high-dimensional global policy learned based on the sampled trajectories generated from the baseline  policy $\beta$ in a supervisory manner by minimizing the KL divergence from the local policy.

The iterative procedure for a general GPS can be summarized as follows: \\
\emph{(Step 1)} Given $\hat{\theta}$, solve 
\[
\begin{split}
\hat{\beta} = \arg \min_{\beta} \ \,  & {\mathbb E}_{\beta}[G({\bm X})] \\
\mbox{s.t.}            \ \, & D_{\rm KL} (\beta({\bm X})\| \pi({\bm X};\hat{\theta})) \leq \epsilon \,.
\end{split}
\]
\emph{(Step 2)} Given $\hat{\beta}$, solve
\[
\begin{split}
\hat{\theta} = \arg \min_{\theta} \ \,  & D_{\rm KL} (\hat{\beta}({\bm X})\| \pi({\bm X};{\theta})) \,.
\end{split}
\]
\emph{(Step 3)} Check convergence and repeat Steps 1 and 2 until a convergence criterion is satisfied.

Various methods have been considered to guide GPS policies. For example, gradient-based local optimal control methods such as DDP and iLQR and sampling-based local approximate optimal control methods such as PI and PI$^{2}$ can be used. Among others, we claim that because of their efficiency in exploration and fast convergence rate, PI-based sampling methods could be more appropriate as guiding policies for GPS. 

\subsection{Model-based reinforcement learning using PI-based policy search and optimization}
\label{sec:disc:mbrl}
In control and robotics, model-based reinforcement learning (MBRL)~\cite{polydoros2017survey,garaffa2021reinforcement,moerland2023model} has been widely investigated because of the potential benefits of data efficiency, effective exploration, and enhanced stability, compared to model-free RL. It is natural to consider methods of path integral control for policy search and optimization in MBRL. The most well-known method of MBRL is the Dyna algorithm~\cite{sutton1990integrated,deisenroth2011pilco} that consists of two iterative learning steps: The first step is to collect data by applying the current policy and learn dynamics model. The second step is to learn or update parameterized policies with data generated from the learned model. The second step alone is known as model-based policy optimization (MBPO)~\cite{janner2019trust,yu2020mopo} that is particularly related to PI-based policy search and optimization.
In other words, parameterized policies discussed in Sections~\ref{sec:PI:alg:feedback}, \ref{sec:PI:alg:PI2}, and \ref{sec:disc:parapolicy} can be improved by using sampling-based path integral control methods for MBRL and MRPO.

\subsection{Sampling efficiency for variance reduction}
\label{sec:disc:sampling}
Let us consider the importance weight defined in\eqref{eq:importance_weight}. Note that if the training policy $\pi$ is optimal, then all simulated trajectories have the same weight. To measure the quality of the sampling strategy, the effective sampling size (ESS) defined as
\begin{equation}
{\rm ESS}^{\pi} = \frac{1}{{\mathbb E}_{{\mathcal P}^{\pi}}\![(\omega^{\pi})^2]}
\end{equation}
measures the variance of the importance weights and can be used to quantify the efficiency of a sampling method~\cite{kotecha2003gaussian,liu2001monte}. A small ESS implies that the associated back-end tasks of estimation or control may result in a large variance. The most important sampling strategies suffer from decreasing ESS over time during prediction. Therefore, quantifying or approximating the ESS of a base-sampling strategy is a major problem in the application of path integral control~\cite{thijssen2015path,zhang2021path}.

\subsection{Extensions to multi-agent path planning}
\label{sec:disc:mapp}
In the literature, there are only a few studies that extend PI control to the stochastic control of multi-agent systems (MASs): path integrals for centralized control~\cite{gomez2016real}, distributed control~\cite{wan2021cooperative,varnai2022multi}, and a two-player zero-sum stochastic differential game (SDG)~\cite{patil2023risk}. A linearly solvable linearly solvable PI control algorithm is proposed for a networked MAS in~\cite{wan2021distributed} and extended to safety-constrained cooperative control of MASs~\cite{song2022generalization,song2023safety} using the control barrier function (CBF) in which the barrier state augmented system is defined to take care of potential conflicts between control objectives and safety requirements. 

In path planning and control for multi-agent systems, it is common to assume that dynamics are independent but costs are interdependent.
Consider the cost-to-go function defined for agent $a$ as 
\[
G_{a,t}^{\bar\pi^{a}} 
= \phi_a(\bar{X}_{a,T}^{\bar\pi_{a}}) + \int_{t}^{T} \!\! L(s, \bar{X}_{a,s}^{\bar\pi_{a}}, {\bar\pi_{a}}(s, \bar{X}_{a,s}^{\bar\pi_{a}})) ds ,
\]
where $\bar\pi_{a}=(\pi_{a}, \pi_{\nu(a)})$ is the joint policy of the ego agent $a$ and its neighbourhood agent ${\nu(a)}$. Similarly the joint state and trajectory are defined as $\bar{X}_{a,s} = (X_{a,s}, X_{\nu(a),s})$ and $\bm{\bar{X}}_{a,t} =(\bar{X}_{a,s})_{s=t}^{T}=(X_{a,s}, X_{\nu(a),s})_{s=t}^{T}$.

As we have observed throughout this paper for single-agent cases, from an algorithmic point of view, the most important computation is to approximate the weights corresponding to the likelihood ratios using MC sampling methods. Similarly, policy updates or improvements in multi-agent systems can be
\[
\pi_{a} \leftarrow \pi_{a} + \sum_{k=1}^{K}  \hat{w}_{a,k}^{\bar{\pi}_{a}} \delta \pi_{a,k}
\]
where the probability weight is defined as
\[
\hat{w}_{a,k}^{\bar{\pi}_{a}} = \frac{\exp\!\left( -\frac{1}{\lambda}G_{a,t}^{(\pi_{a,k}, \pi_{\nu(a)})} \right)}{\sum_{\kappa=1}^{K}\exp\!\left( -\frac{1}{\lambda}G_{a,t}^{(\pi_{a,\kappa}, \pi_{\nu(a)})} \right)} 
\]
for which randomly perturbed policies $\pi_{a,k}= \pi_{a} + \delta \pi_{a,k}$ are used to simulate the controlled trajectories and compute the associated costs $G_{a,t}^{(\pi_{a,k}, \pi_{\nu(a)})}$. Here, the learning process is assumed to be asynchronous, in the sense that the policies of the neighborhood agents $\nu(a)$ are fixed when updating the policy for agent $a$ in accordance with the simulation of the augmented trajectories $\bm{\bar X}_{a}$.
Here, the individual agent's policy can be either MPC-like open-loop (feedforward) control inputs or parameterized (deterministic or stochastic) state-feedback controllers.

\subsection{MPPI for trajectory optimization on manifolds}
\label{sec:disc:manifold}
Trajectory optimization using differential geometry is very common in robotics and has been studied under manifolds such as special orthogonal and Euclidean groups ${\rm SO}(3)$ and ${\rm SE}(3)$~\cite{watterson2018trajectory,bonalli2019trajectory,watterson2020trajectory,osa2022motion}.
In~\cite{bonalli2019trajectory}, gradient-based sequential convex programming on manifolds was used for trajectory optimization. It was expected that many theoretical and computational frameworks for the optimization of manifolds~\cite{boumal2014manopt,boumal2023introduction} could be applied to robotic trajectory optimization.

Applying methods of sampling-based path integral control such as MPPI to trajectory optimization on manifolds is not trivial because it requires effective accelerated approaches to generate the sampled trajectories on manifolds and sampling trajectories on manifolds with kinematic constraints are not trivial. Thus, one could employ the methods used for unscented {K}alman filtering on manifolds {(UKF-M)}~\cite{menegaz2018unscented,brossard2020code,li2020unscented,cantelobre2020real}. We leave this research topic of sampling-based path integral control for trajectory optimization on manifolds for potential future work. 

\subsection{Motion planning for mobile robots and manipulators}
\label{sec:disc:mobile-manipulator}
While policy parameterization discussed in Section~\ref{sec:disc:parapolicy} has shown great success in robotic manipulators and locomotion control, policy parameterization is not trivial and even not appropriate for path planning or trajectory optimization for autonomous mobile robots. This is because optimal trajectory should be determined by considering the relative pose of the robot with respect to the obstacles and the goal pose as well as the robot's ego-pose. Such relative poses can be encoded into the associated optimal control problem (OCP) as constraints and feasibility of the OCP with a parameterized policy is hard to be guaranteed for mobile robot dynamic environment. This brings us a conclusion that it is more appropriate to use the MPC-like open-loop control inputs for trajectory optimization of mobile robots, which implies that MPPI and its variations would become more successful with autonomous mobile robot navigation than PI$^2$-based policy search and optimization.

\section{Conclusions}\label{sec:conclusion}
In this paper, we present an overview of the fundamental theoretical developments and recent advances in path integral control with a focus on sampling-based stochastic trajectory optimization. The theoretical and algorithmic frameworks of several optimal control methods employing the path integral control framework are provided, and their similarities and differences are reviewed. Python, MATLAB and ROS2/Gazebo simulation results are provided to demonstrate the effectiveness of various path integral control methods. Discussions on policy parameterization and optimization in policy search adopting path integral control, connections to model-based reinforcement learning, efficiency of sampling strategies, extending the path integral control framework to multi-agent optimal control problems, and path integral control for the trajectory optimization of manifolds are provided. We expect that sampling-based stochastic trajectory optimization employing path integral control can be applied to practical engineering problems, particularly for agile mobile robot navigation and control.

\bibliographystyle{IEEEtran}
\bibliography{pathintegralcontrol_tutorial}

\begin{thebibliography}{100}
\providecommand{\url}[1]{#1}
\csname url@samestyle\endcsname
\providecommand{\newblock}{\relax}
\providecommand{\bibinfo}[2]{#2}
\providecommand{\BIBentrySTDinterwordspacing}{\spaceskip=0pt\relax}
\providecommand{\BIBentryALTinterwordstretchfactor}{4}
\providecommand{\BIBentryALTinterwordspacing}{\spaceskip=\fontdimen2\font plus
\BIBentryALTinterwordstretchfactor\fontdimen3\font minus
  \fontdimen4\font\relax}
\providecommand{\BIBforeignlanguage}[2]{{%
\expandafter\ifx\csname l@#1\endcsname\relax
\typeout{** WARNING: IEEEtran.bst: No hyphenation pattern has been}%
\typeout{** loaded for the language `#1'. Using the pattern for}%
\typeout{** the default language instead.}%
\else
\language=\csname l@#1\endcsname
\fi
#2}}
\providecommand{\BIBdecl}{\relax}
\BIBdecl

\bibitem{von1992direct}
O.~Von~Stryk and R.~Bulirsch, ``Direct and indirect methods for trajectory
  optimization,'' \emph{Annals of operations research}, vol.~37, no.~1, pp.
  357--373, 1992.

\bibitem{betts1998survey}
J.~T. Betts, ``Survey of numerical methods for trajectory optimization,''
  \emph{Journal of guidance, control, and dynamics}, vol.~21, no.~2, pp.
  193--207, 1998.

\bibitem{rao2014trajectory}
A.~V. Rao, ``Trajectory optimization: {A} survey,'' \emph{Optimization and
  optimal control in automotive systems}, pp. 3--21, 2014.

\bibitem{choset2005principles}
H.~Choset, K.~M. Lynch, S.~Hutchinson, G.~A. Kantor, and W.~Burgard,
  \emph{Principles of Robot Motion: Theory, Algorithms, and
  Implementations}.\hskip 1em plus 0.5em minus 0.4em\relax Cambridge,
  Massachusetts: MIT Press, 2005.

\bibitem{latombe2012robot}
J.-C. Latombe, \emph{Robot Motion Planning}.\hskip 1em plus 0.5em minus
  0.4em\relax New York: Springer Science \& Business Media, 2012, vol. 124.

\bibitem{lavalle2006planning}
S.~M. LaValle, \emph{Planning Algorithms}.\hskip 1em plus 0.5em minus
  0.4em\relax New York: Cambridge University Press, 2006.

\bibitem{paden2016survey}
B.~Paden, M.~{\v{C}}{\'a}p, S.~Z. Yong, D.~Yershov, and E.~Frazzoli, ``A survey
  of motion planning and control techniques for self-driving urban vehicles,''
  \emph{IEEE Transactions on Intelligent Vehicles}, vol.~1, no.~1, pp. 33--55,
  2016.

\bibitem{claussmann2019review}
L.~Claussmann, M.~Revilloud, D.~Gruyer, and S.~Glaser, ``A review of motion
  planning for highway autonomous driving,'' \emph{IEEE Transactions on
  Intelligent Transportation Systems}, vol.~21, no.~5, pp. 1826--1848, 2019.

\bibitem{teng2023motion}
S.~Teng, X.~Hu, P.~Deng, B.~Li, Y.~Li, Y.~Ai, D.~Yang, L.~Li, Z.~Xuanyuan,
  F.~Zhu \emph{et~al.}, ``Motion planning for autonomous driving: The state of
  the art and future perspectives,'' \emph{IEEE Transactions on Intelligent
  Vehicles}, 2023.

\bibitem{song2021autonomous}
Y.~Song, M.~Steinweg, E.~Kaufmann, and D.~Scaramuzza, ``Autonomous drone racing
  with deep reinforcement learning,'' in \emph{2021 IEEE/RSJ International
  Conference on Intelligent Robots and Systems (IROS)}.\hskip 1em plus 0.5em
  minus 0.4em\relax IEEE, 2021, pp. 1205--1212.

\bibitem{han2021fast}
Z.~Han, Z.~Wang, N.~Pan, Y.~Lin, C.~Xu, and F.~Gao, ``{Fast-Racing}: {A}n
  open-source strong baseline for $\mathrm se$(3) planning in autonomous drone
  racing,'' \emph{IEEE Robotics and Automation Letters}, vol.~6, no.~4, pp.
  8631--8638, 2021.

\bibitem{hanover2023autonomous}
D.~Hanover, A.~Loquercio, L.~Bauersfeld, A.~Romero, R.~Penicka, Y.~Song,
  G.~Cioffi, E.~Kaufmann, and D.~Scaramuzza, ``Autonomous drone racing: {A}
  survey,'' \emph{arXiv e-prints, pp. arXiv--2301}, 2023.

\bibitem{lan2021survey}
M.~Lan, S.~Lai, T.~H. Lee, and B.~M. Chen, ``A survey of motion and task
  planning techniques for unmanned multicopter systems,'' \emph{Unmanned
  Systems}, vol.~9, no.~02, pp. 165--198, 2021.

\bibitem{wang2021trajectory}
M.~Wang, J.~Diepolder, S.~Zhang, M.~S{\"o}pper, and F.~Holzapfel, ``Trajectory
  optimization-based maneuverability assessment of evtol aircraft,''
  \emph{Aerospace Science and Technology}, vol. 117, p. 106903, 2021.

\bibitem{park2023trajectory}
J.~Park, I.~Kim, J.~Suk, and S.~Kim, ``Trajectory optimization for takeoff and
  landing phase of uam considering energy and safety,'' \emph{Aerospace Science
  and Technology}, vol. 140, p. 108489, 2023.

\bibitem{pradeep2020wind}
P.~Pradeep, T.~A. Lauderdale, G.~B. Chatterji, K.~Sheth, C.~F. Lai, B.~Sridhar,
  K.-M. Edholm, and H.~Erzberger, ``Wind-optimal trajectories for multirotor
  evtol aircraft on uam missions,'' in \emph{Aiaa Aviation 2020 Forum}, 2020,
  p. 3271.

\bibitem{kwon2020convex}
H.-H. Kwon and H.-L. Choi, ``A convex programming approach to mid-course
  trajectory optimization for air-to-ground missiles,'' \emph{International
  Journal of Aeronautical and Space Sciences}, vol.~21, pp. 479--492, 2020.

\bibitem{roh2020l1}
H.~Roh, Y.-J. Oh, M.-J. Tahk, K.-J. Kwon, and H.-H. Kwon, ``L1 penalized
  sequential convex programming for fast trajectory optimization: With
  application to optimal missile guidance,'' \emph{International Journal of
  Aeronautical and Space Sciences}, vol.~21, pp. 493--503, 2020.

\bibitem{garcia2005trajectory}
I.~Garcia and J.~P. How, ``Trajectory optimization for satellite
  reconfiguration maneuvers with position and attitude constraints,'' in
  \emph{Proceedings of the 2005, American Control Conference, 2005.}\hskip 1em
  plus 0.5em minus 0.4em\relax IEEE, 2005, pp. 889--894.

\bibitem{weiss2014spacecraft}
A.~Weiss, F.~Leve, M.~Baldwin, J.~R. Forbes, and I.~Kolmanovsky, ``Spacecraft
  constrained attitude control using positively invariant constraint admissible
  sets on {SO}(3)$\times\mathbb{R}^3$,'' in \emph{2014 American Control
  Conference}.\hskip 1em plus 0.5em minus 0.4em\relax IEEE, 2014, pp.
  4955--4960.

\bibitem{gatherer2019magnetorquer}
A.~Gatherer and Z.~Manchester, ``Magnetorquer-only attitude control of small
  satellites using trajectory optimization,'' in \emph{Proceedings of AAS/AIAA
  Astrodynamics Specialist Conference}, 2019.

\bibitem{malyuta2021advances}
D.~Malyuta, Y.~Yu, P.~Elango, and B.~A{\c{c}}{\i}kme{\c{s}}e, ``Advances in
  trajectory optimization for space vehicle control,'' \emph{Annual Reviews in
  Control}, vol.~52, pp. 282--315, 2021.

\bibitem{dearing2022efficient}
T.~L. Dearing, J.~Hauser, X.~Chen, M.~M. Nicotra, and C.~Petersen, ``Efficient
  trajectory optimization for constrained spacecraft attitude maneuvers,''
  \emph{Journal of Guidance, Control, and Dynamics}, vol.~45, no.~4, pp.
  638--650, 2022.

\bibitem{canny1988complexity}
J.~Canny, \emph{The Complexity of Robot Motion Planning}.\hskip 1em plus 0.5em
  minus 0.4em\relax Cambridge, Massachusetts: MIT Press, 1988.

\bibitem{elbanhawi2014sampling}
M.~Elbanhawi and M.~Simic, ``Sampling-based robot motion planning: {A}
  review,'' \emph{IEEE Access}, vol.~2, pp. 56--77, 2014.

\bibitem{kingston2018sampling}
Z.~Kingston, M.~Moll, and L.~E. Kavraki, ``Sampling-based methods for motion
  planning with constraints,'' \emph{Annual Review of Control, Robotics, and
  Autonomous Systems}, vol.~1, no.~1, pp. 159--185, 2018.

\bibitem{betts2010practical}
J.~T. Betts, \emph{Practical Methods for Optimal Control and Estimation Using
  Nonlinear Programming}.\hskip 1em plus 0.5em minus 0.4em\relax Philadelphia,
  PA: SIAM, 2010.

\bibitem{kelly2017introduction}
M.~Kelly, ``An introduction to trajectory optimization: {H}ow to do your own
  direct collocation,'' \emph{SIAM Review}, vol.~59, no.~4, pp. 849--904, 2017.

\bibitem{bonalli2019gusto}
R.~Bonalli, A.~Cauligi, A.~Bylard, and M.~Pavone, ``{GuSTO}: {G}uaranteed
  sequential trajectory optimization via sequential convex programming,'' in
  \emph{2019 International conference on robotics and automation (ICRA)}.\hskip
  1em plus 0.5em minus 0.4em\relax IEEE, 2019, pp. 6741--6747.

\bibitem{howell2019altro}
T.~A. Howell, B.~E. Jackson, and Z.~Manchester, ``{ALTRO}: {A} fast solver for
  constrained trajectory optimization,'' in \emph{2019 IEEE/RSJ International
  Conference on Intelligent Robots and Systems (IROS)}.\hskip 1em plus 0.5em
  minus 0.4em\relax IEEE, 2019, pp. 7674--7679.

\bibitem{manyam2021trajectory}
S.~G. Manyam, D.~W. Casbeer, I.~E. Weintraub, and C.~Taylor, ``Trajectory
  optimization for rendezvous planning using quadratic {B}{\'e}zier curves,''
  in \emph{2021 IEEE/RSJ International Conference on Intelligent Robots and
  Systems (IROS)}.\hskip 1em plus 0.5em minus 0.4em\relax IEEE, 2021, pp.
  1405--1412.

\bibitem{malyuta2022convex}
D.~Malyuta, T.~P. Reynolds, M.~Szmuk, T.~Lew, R.~Bonalli, M.~Pavone, and
  B.~A{\c{c}}{\i}kme{\c{s}}e, ``< convex optimization for trajectory
  generation: {A} tutorial on generating dynamically feasible trajectories
  reliably and efficiently,'' \emph{IEEE Control Systems Magazine}, vol.~42,
  no.~5, pp. 40--113, 2022.

\bibitem{jacobson1970differential}
D.~H. Jacobson and D.~Q. Mayne, \emph{Differential Dynamic Programming}.\hskip
  1em plus 0.5em minus 0.4em\relax new York: Elsevier Publishing Company, 1970,
  no.~24.

\bibitem{mayne1973differential}
D.~Q. Mayne, ``Differential dynamic programming--a unified approach to the
  optimization of dynamic systems,'' in \emph{Control and dynamic
  systems}.\hskip 1em plus 0.5em minus 0.4em\relax Elsevier, 1973, vol.~10, pp.
  179--254.

\bibitem{xie2017differential}
Z.~Xie, C.~K. Liu, and K.~Hauser, ``Differential dynamic programming with
  nonlinear constraints,'' in \emph{2017 IEEE International Conference on
  Robotics and Automation (ICRA)}.\hskip 1em plus 0.5em minus 0.4em\relax IEEE,
  2017, pp. 695--702.

\bibitem{chen2019autonomous}
J.~Chen, W.~Zhan, and M.~Tomizuka, ``Autonomous driving motion planning with
  constrained iterative {LQR},'' \emph{IEEE Transactions on Intelligent
  Vehicles}, vol.~4, no.~2, pp. 244--254, 2019.

\bibitem{pavlov2021interior}
A.~Pavlov, I.~Shames, and C.~Manzie, ``Interior point differential dynamic
  programming,'' \emph{IEEE Transactions on Control Systems Technology},
  vol.~29, no.~6, pp. 2720--2727, 2021.

\bibitem{cao2022direct}
K.~Cao, M.~Cao, S.~Yuan, and L.~Xie, ``{DIRECT:} {A} differential dynamic
  programming based framework for trajectory generation,'' \emph{IEEE Robotics
  and Automation Letters}, vol.~7, no.~2, pp. 2439--2446, 2022.

\bibitem{chatzinikolaidis2021trajectory}
I.~Chatzinikolaidis and Z.~Li, ``Trajectory optimization of contact-rich
  motions using implicit differential dynamic programming,'' \emph{IEEE
  Robotics and Automation Letters}, vol.~6, no.~2, pp. 2626--2633, 2021.

\bibitem{kim2022extension}
M.-G. Kim and K.-K.~K. Kim, ``An extension of interior point differential
  dynamic programming for optimal control problems with second-order conic
  constraints,'' \emph{Transactions of the Korean Institute of Electrical
  Engineers}, vol.~71, no.~11, pp. 1666--1672, 2022.

\bibitem{zhong2020hybrid}
X.~Zhong, J.~Tian, H.~Hu, and X.~Peng, ``Hybrid path planning based on safe a*
  algorithm and adaptive window approach for mobile robot in large-scale
  dynamic environment,'' \emph{Journal of Intelligent \& Robotic Systems},
  vol.~99, no.~1, pp. 65--77, 2020.

\bibitem{sun2022comparative}
S.~Sun, A.~Romero, P.~Foehn, E.~Kaufmann, and D.~Scaramuzza, ``A comparative
  study of nonlinear {MPC} and differential-flatness-based control for
  quadrotor agile flight,'' \emph{IEEE Transactions on Robotics}, vol.~38,
  no.~6, pp. 3357--3373, 2022.

\bibitem{faessler2017differential}
M.~Faessler, A.~Franchi, and D.~Scaramuzza, ``Differential flatness of
  quadrotor dynamics subject to rotor drag for accurate tracking of high-speed
  trajectories,'' \emph{IEEE Robotics and Automation Letters}, vol.~3, no.~2,
  pp. 620--626, 2017.

\bibitem{FORCESPro}
A.~Domahidi and J.~Jerez, ``{FORCES Professional},'' EmbotechAG,
  url=https://embotech.com/FORCES-Pro, 2014--2023.

\bibitem{gammell2020asymptotically}
J.~D. Gammell and M.~P. Strub, ``Asymptotically optimal sampling-based motion
  planning methods,'' \emph{arXiv preprint arXiv:2009.10484}, 2020.

\bibitem{kuffner2000rrt}
J.~J. Kuffner and S.~M. LaValle, ``{RRT}-connect: {A}n efficient approach to
  single-query path planning,'' in \emph{2000 IEEE International Conference on
  Robotics and Automation (ICRA)}, vol.~2.\hskip 1em plus 0.5em minus
  0.4em\relax IEEE, 2000, pp. 995--1001.

\bibitem{janson2018deterministic}
L.~Janson, B.~Ichter, and M.~Pavone, ``Deterministic sampling-based motion
  planning: Optimality, complexity, and performance,'' \emph{The International
  Journal of Robotics Research}, vol.~37, no.~1, pp. 46--61, 2018.

\bibitem{campos2017hybrid}
L.~Campos-Mac{\'\i}as, D.~G{\'o}mez-Guti{\'e}rrez, R.~Aldana-L{\'o}pez,
  R.~de~la Guardia, and J.~I. Parra-Vilchis, ``A hybrid method for online
  trajectory planning of mobile robots in cluttered environments,'' \emph{IEEE
  Robotics and Automation Letters}, vol.~2, no.~2, pp. 935--942, 2017.

\bibitem{ravankar2020hpprm}
A.~A. Ravankar, A.~Ravankar, T.~Emaru, and Y.~Kobayashi, ``{HPPRM}: {H}ybrid
  potential based probabilistic roadmap algorithm for improved dynamic path
  planning of mobile robots,'' \emph{IEEE Access}, vol.~8, pp.
  221\,743--221\,766, 2020.

\bibitem{kiani2021adapted}
F.~Kiani, A.~Seyyedabbasi, R.~Aliyev, M.~U. Gulle, H.~Basyildiz, and M.~A.
  Shah, ``{Adapted-RRT}: {N}ovel hybrid method to solve three-dimensional path
  planning problem using sampling and metaheuristic-based algorithms,''
  \emph{Neural Computing and Applications}, vol.~33, no.~22, pp.
  15\,569--15\,599, 2021.

\bibitem{yu2022novel}
Z.~Yu, Z.~Si, X.~Li, D.~Wang, and H.~Song, ``A novel hybrid particle swarm
  optimization algorithm for path planning of uavs,'' \emph{IEEE Internet of
  Things Journal}, vol.~9, no.~22, pp. 22\,547--22\,558, 2022.

\bibitem{sucan2012open}
I.~A. Sucan, M.~Moll, and L.~E. Kavraki, ``The open motion planning library,''
  \emph{IEEE Robotics \& Automation Magazine}, vol.~19, no.~4, pp. 72--82,
  2012.

\bibitem{kalakrishnan2011stomp}
M.~Kalakrishnan, S.~Chitta, E.~Theodorou, P.~Pastor, and S.~Schaal, ``{STOMP}:
  {S}tochastic trajectory optimization for motion planning,'' in \emph{IEEE
  International Conference on Robotics and Automation (ICRA)}.\hskip 1em plus
  0.5em minus 0.4em\relax IEEE, 2011, pp. 4569--4574.

\bibitem{likhachev2010search}
\BIBentryALTinterwordspacing
M.~Likhachev, ``Search-based planning lab,'' 2010. [Online]. Available:
  \url{http://sbpl.net/Home}
\BIBentrySTDinterwordspacing

\bibitem{zucker2013chomp}
M.~Zucker, N.~Ratliff, A.~D. Dragan, M.~Pivtoraiko, M.~Klingensmith, C.~M.
  Dellin, J.~A. Bagnell, and S.~S. Srinivasa, ``Chomp: Covariant hamiltonian
  optimization for motion planning,'' \emph{The International Journal of
  Robotics Research}, vol.~32, no. 9-10, pp. 1164--1193, 2013.

\bibitem{kappen2005linear}
H.~J. Kappen, ``Linear theory for control of nonlinear stochastic systems,''
  \emph{Physical review letters}, vol.~95, no.~20, p. 200201, 2005.

\bibitem{rubinstein2004cross}
R.~Y. Rubinstein and D.~P. Kroese, \emph{The Cross-Entropy Method: A Unified
  Approach to Combinatorial Optimization, Monte-Carlo Simulation and Machine
  Learning}.\hskip 1em plus 0.5em minus 0.4em\relax New York: Springer-Verlag,
  2004, vol. 133.

\bibitem{gomez2014policy}
V.~G{\'o}mez, H.~J. Kappen, J.~Peters, and G.~Neumann, ``Policy search for path
  integral control,'' in \emph{Joint European Conference on Machine Learning
  and Knowledge Discovery in Databases}.\hskip 1em plus 0.5em minus 0.4em\relax
  Springer, 2014, pp. 482--497.

\bibitem{williams2017model}
G.~Williams, A.~Aldrich, and E.~A. Theodorou, ``Model predictive path integral
  control: {F}rom theory to parallel computation,'' \emph{Journal of Guidance,
  Control, and Dynamics}, vol.~40, no.~2, pp. 344--357, 2017.

\bibitem{rubinstein2016simulation}
R.~Y. Rubinstein and D.~P. Kroese, \emph{Simulation and the {M}onte {C}arlo
  Method}, 2nd~ed.\hskip 1em plus 0.5em minus 0.4em\relax New York: John Wiley
  \& Sons, 2016.

\bibitem{thijssen2018consistent}
S.~Thijssen and H.~Kappen, ``Consistent adaptive multiple importance sampling
  and controlled diffusions,'' \emph{arXiv preprint arXiv:1803.07966}, 2018.

\bibitem{williams2016aggressive}
G.~Williams, P.~Drews, B.~Goldfain, J.~M. Rehg, and E.~A. Theodorou,
  ``Aggressive driving with model predictive path integral control,'' in
  \emph{IEEE International Conference on Robotics and Automation (ICRA)}.\hskip
  1em plus 0.5em minus 0.4em\relax IEEE, 2016, pp. 1433--1440.

\bibitem{williams2018information}
------, ``Information-theoretic model predictive control: {T}heory and
  applications to autonomous driving,'' \emph{IEEE Transactions on Robotics},
  vol.~34, no.~6, pp. 1603--1622, 2018.

\bibitem{williams2019mppi}
G.~R. Williams, ``Model predictive path integral control: {T}heoretical
  foundations and applications to autonomous driving,'' Ph.D. dissertation,
  Georgia Institute of Techonology, 2019.

\bibitem{pan2015sample}
Y.~Pan, E.~Theodorou, and M.~Kontitsis, ``Sample efficient path integral
  control under uncertainty,'' \emph{Advances in Neural Information Processing
  Systems}, vol.~28, 2015.

\bibitem{abraham2020model}
I.~Abraham, A.~Handa, N.~Ratliff, K.~Lowrey, T.~D. Murphey, and D.~Fox,
  ``Model-based generalization under parameter uncertainty using path integral
  control,'' \emph{IEEE Robotics and Automation Letters}, vol.~5, no.~2, pp.
  2864--2871, 2020.

\bibitem{pravitra2020}
J.~Pravitra, K.~A. Ackerman, C.~Cao, N.~Hovakimyan, and E.~A. Theodorou,
  ``{L1-Adaptive MPPI} architecture for robust and agile control of
  multirotors,'' in \emph{2020 IEEE/RSJ International Conference on Intelligent
  Robots and Systems (IROS)}, 2020, pp. 7661--7666.

\bibitem{williams2015gpu}
G.~Williams, E.~Rombokas, and T.~Daniel, ``{GPU} based path integral control
  with learned dynamics,'' \emph{arXiv preprint arXiv:1503.00330}, 2015.

\bibitem{okada2017path}
M.~Okada, T.~Aoshima, and L.~Rigazio, ``Path integral networks: {E}nd-to-end
  differentiable optimal control,'' \emph{arXiv preprint arXiv:1706.09597},
  2017.

\bibitem{mohamed2023gp}
I.~S. Mohamed, M.~Ali, and L.~Liu, ``{GP}-guided {MPPI} for efficient
  navigation in complex unknown cluttered environments,'' \emph{arXiv preprint
  arXiv:2307.04019}, 2023.

\bibitem{gandhi2021robust}
M.~S. Gandhi, B.~Vlahov, J.~Gibson, G.~Williams, and E.~A. Theodorou, ``Robust
  model predictive path integral control: {A}nalysis and performance
  guarantees,'' \emph{IEEE Robotics and Automation Letters}, vol.~6, no.~2, pp.
  1423--1430, 2021.

\bibitem{arruda2017uncertainty}
E.~Arruda, M.~J. Mathew, M.~Kopicki, M.~Mistry, M.~Azad, and J.~L. Wyatt,
  ``Uncertainty averse pushing with model predictive path integral control,''
  in \emph{2017 IEEE-RAS 17th International Conference on Humanoid Robotics
  (Humanoids)}.\hskip 1em plus 0.5em minus 0.4em\relax IEEE, 2017, pp.
  497--502.

\bibitem{raisi2022fault}
M.~Raisi, A.~Noohian, and S.~Fallah, ``A fault-tolerant and robust controller
  using model predictive path integral control for free-flying space robots,''
  \emph{Frontiers in Robotics and AI}, vol.~9, p. 1027918, 2022.

\bibitem{zeng2021safety}
J.~Zeng, B.~Zhang, and K.~Sreenath, ``Safety-critical model predictive control
  with discrete-time control barrier function,'' in \emph{2021 American Control
  Conference (ACC)}.\hskip 1em plus 0.5em minus 0.4em\relax IEEE, 2021, pp.
  3882--3889.

\bibitem{tao2022path}
C.~Tao, H.-J. Yoon, H.~Kim, N.~Hovakimyan, and P.~Voulgaris, ``Path integral
  methods with stochastic control barrier functions,'' in \emph{2022 IEEE 61st
  Conference on Decision and Control (CDC)}.\hskip 1em plus 0.5em minus
  0.4em\relax IEEE, 2022, pp. 1654--1659.

\bibitem{yin2023shield}
J.~Yin, C.~Dawson, C.~Fan, and P.~Tsiotras, ``Shield model predictive path
  integral: {A} computationally efficient robust {MPC} approach using control
  barrier functions,'' \emph{arXiv preprint arXiv:2302.11719}, 2023.

\bibitem{yin2022trajectory}
J.~Yin, Z.~Zhang, E.~Theodorou, and P.~Tsiotras, ``Trajectory distribution
  control for model predictive path integral control using covariance
  steering,'' in \emph{2022 International Conference on Robotics and Automation
  (ICRA)}.\hskip 1em plus 0.5em minus 0.4em\relax IEEE, 2022, pp. 1478--1484.

\bibitem{barbosa2021risk}
F.~S. Barbosa, B.~Lacerda, P.~Duckworth, J.~Tumova, and N.~Hawes, ``Risk-aware
  motion planning in partially known environments,'' in \emph{2021 60th IEEE
  Conference on Decision and Control (CDC)}.\hskip 1em plus 0.5em minus
  0.4em\relax IEEE, 2021, pp. 5220--5226.

\bibitem{wang2021adaptive}
Z.~Wang, O.~So, K.~Lee, and E.~A. Theodorou, ``Adaptive risk sensitive model
  predictive control with stochastic search,'' in \emph{Learning for Dynamics
  and Control}.\hskip 1em plus 0.5em minus 0.4em\relax PMLR, 2021, pp.
  510--522.

\bibitem{cai2022probabilistic}
X.~Cai, M.~Everett, L.~Sharma, P.~R. Osteen, and J.~P. How, ``Probabilistic
  traversability model for risk-aware motion planning in off-road
  environments,'' \emph{arXiv preprint arXiv:2210.00153}, 2022.

\bibitem{yin2023risk}
J.~Yin, Z.~Zhang, and P.~Tsiotras, ``Risk-aware model predictive path integral
  control using conditional value-at-risk,'' in \emph{2023 IEEE International
  Conference on Robotics and Automation (ICRA)}.\hskip 1em plus 0.5em minus
  0.4em\relax IEEE, 2023, pp. 7937--7943.

\bibitem{tao2023rrt}
C.~Tao, H.~Kim, and N.~Hovakimyan, ``{RRT} guided model predictive path
  integral method,'' \emph{arXiv preprint arXiv:2301.13143}, 2023.

\bibitem{theodorou2012relative}
E.~A. Theodorou and E.~Todorov, ``Relative entropy and free energy dualities:
  {C}onnections to path integral and {KL} control,'' in \emph{2012 51st IEEE
  Conference on Decision and Control (CDC)}.\hskip 1em plus 0.5em minus
  0.4em\relax IEEE, 2012, pp. 1466--1473.

\bibitem{ijspeert2002learning}
A.~Ijspeert, J.~Nakanishi, and S.~Schaal, ``Learning attractor landscapes for
  learning motor primitives,'' \emph{Advances in neural information processing
  systems}, vol.~15, 2002.

\bibitem{theodorou2010generalized}
E.~Theodorou, J.~Buchli, and S.~Schaal, ``A generalized path integral control
  approach to reinforcement learning,'' \emph{The Journal of Machine Learning
  Research}, vol.~11, pp. 3137--3181, 2010.

\bibitem{levine2013guided}
S.~Levine and V.~Koltun, ``Guided policy search,'' in \emph{International
  Conference on Machine Learning}.\hskip 1em plus 0.5em minus 0.4em\relax PMLR,
  2013, pp. 1--9.

\bibitem{montgomery2016guided}
W.~H. Montgomery and S.~Levine, ``Guided policy search via approximate mirror
  descent,'' \emph{Advances in Neural Information Processing Systems (NIPS)},
  vol.~29, 2016.

\bibitem{kim2022mppi}
M.-G. Kim and K.-K.~K. Kim, ``{MPPI-IPDDP}: {H}ybrid method of collision-free
  smooth trajectory generation for autonomous robots,'' \emph{arXiv preprint
  arXiv:2208.02439}, 2022.

\bibitem{thalmeier2020adaptive}
D.~Thalmeier, H.~J. Kappen, S.~Totaro, and V.~G{\'o}mez, ``Adaptive smoothing
  for path integral control,'' \emph{The Journal of Machine Learning Research},
  vol.~21, no.~1, pp. 7814--7850, 2020.

\bibitem{sarkka2008unscented}
S.~S{\"a}rkk{\"a}, ``Unscented rauch--tung--striebel smoother,'' \emph{IEEE
  transactions on automatic control}, vol.~53, no.~3, pp. 845--849, 2008.

\bibitem{ruiz2017particle}
H.-C. Ruiz and H.~J. Kappen, ``Particle smoothing for hidden diffusion
  processes: Adaptive path integral smoother,'' \emph{IEEE Transactions on
  Signal Processing}, vol.~65, no.~12, pp. 3191--3203, 2017.

\bibitem{neve2022comparative}
T.~Neve, T.~Lefebvre, and G.~Crevecoeur, ``Comparative study of sample based
  model predictive control with application to autonomous racing,'' in
  \emph{2022 IEEE/ASME International Conference on Advanced Intelligent
  Mechatronics (AIM)}.\hskip 1em plus 0.5em minus 0.4em\relax IEEE, 2022, pp.
  1632--1638.

\bibitem{kim2022smooth}
T.~Kim, G.~Park, K.~Kwak, J.~Bae, and W.~Lee, ``Smooth model predictive path
  integral control without smoothing,'' \emph{IEEE Robotics and Automation
  Letters}, vol.~7, no.~4, pp. 10\,406--10\,413, 2022.

\bibitem{lefebvre2022entropy}
T.~Lefebvre and G.~Crevecoeur, ``Entropy regularised deterministic optimal
  control: {F}rom path integral solution to sample-based trajectory
  optimisation,'' in \emph{2022 IEEE/ASME International Conference on Advanced
  Intelligent Mechatronics (AIM)}.\hskip 1em plus 0.5em minus 0.4em\relax IEEE,
  2022, pp. 401--408.

\bibitem{lambert2020stein}
A.~Lambert, A.~Fishman, D.~Fox, B.~Boots, and F.~Ramos, ``Stein variational
  model predictive control,'' \emph{arXiv preprint arXiv:2011.07641}, 2020.

\bibitem{eysenbach2019if}
B.~Eysenbach and S.~Levine, ``If {MaxEnt RL} is the answer, what is the
  question?'' \emph{arXiv preprint arXiv:1910.01913}, 2019.

\bibitem{whittle1991likelihood}
P.~Whittle, ``Likelihood and cost as path integrals,'' \emph{Journal of the
  Royal Statistical Society: Series B (Methodological)}, vol.~53, no.~3, pp.
  505--529, 1991.

\bibitem{kappen2012optimal}
H.~J. Kappen, V.~G{\'o}mez, and M.~Opper, ``Optimal control as a graphical
  model inference problem,'' \emph{Machine learning}, vol.~87, no.~2, pp.
  159--182, 2012.

\bibitem{watson2020stochastic}
J.~Watson, H.~Abdulsamad, and J.~Peters, ``Stochastic optimal control as
  approximate input inference,'' in \emph{Conference on Robot Learning}.\hskip
  1em plus 0.5em minus 0.4em\relax PMLR, 2020, pp. 697--716.

\bibitem{haarnoja2017reinforcement}
T.~Haarnoja, H.~Tang, P.~Abbeel, and S.~Levine, ``Reinforcement learning with
  deep energy-based policies,'' in \emph{International conference on machine
  learning}.\hskip 1em plus 0.5em minus 0.4em\relax PMLR, 2017, pp. 1352--1361.

\bibitem{levine2018reinforcement}
S.~Levine, ``Reinforcement learning and control as probabilistic inference:
  {T}utorial and review,'' \emph{arXiv preprint arXiv:1805.00909}, 2018.

\bibitem{martino2015adaptive}
L.~Martino, V.~Elvira, D.~Luengo, and J.~Corander, ``An adaptive population
  importance sampler: {L}earning from uncertainty,'' \emph{IEEE Transactions on
  Signal Processing}, vol.~63, no.~16, pp. 4422--4437, 2015.

\bibitem{stich2017safe}
S.~U. Stich, A.~Raj, and M.~Jaggi, ``Safe adaptive importance sampling,''
  \emph{Advances in Neural Information Processing Systems}, vol.~30, 2017.

\bibitem{bugallo2017adaptive}
M.~F. Bugallo, V.~Elvira, L.~Martino, D.~Luengo, J.~Miguez, and P.~M. Djuric,
  ``Adaptive importance sampling: {T}he past, the present, and the future,''
  \emph{IEEE Signal Processing Magazine}, vol.~34, no.~4, pp. 60--79, 2017.

\bibitem{kappen2016adaptive}
H.~J. Kappen and H.~C. Ruiz, ``Adaptive importance sampling for control and
  inference,'' \emph{Journal of Statistical Physics}, vol. 162, no.~5, pp.
  1244--1266, 2016.

\bibitem{asmar2023model}
D.~M. Asmar, R.~Senanayake, S.~Manuel, and M.~J. Kochenderfer, ``Model
  predictive optimized path integral strategies,'' in \emph{2023 IEEE
  International Conference on Robotics and Automation (ICRA)}, 2023, pp.
  3182--3188.

\bibitem{carius2022constrained}
J.~Carius, R.~Ranftl, F.~Farshidian, and M.~Hutter, ``Constrained stochastic
  optimal control with learned importance sampling: A path integral approach,''
  \emph{The International Journal of Robotics Research}, vol.~41, no.~2, pp.
  189--209, 2022.

\bibitem{Arouna+2004+1+24}
B.~Arouna, ``Adaptative {Monte Carlo} method, a variance reduction technique,''
  \emph{Monte Carlo Methods and Applications}, vol.~10, no.~1, pp. 1--24, 2004.

\bibitem{de2005tutorial}
P.-T. De~Boer, D.~P. Kroese, S.~Mannor, and R.~Y. Rubinstein, ``A tutorial on
  the cross-entropy method,'' \emph{Annals of operations research}, vol. 134,
  no.~1, pp. 19--67, 2005.

\bibitem{kobilarov2012cross}
M.~Kobilarov, ``Cross-entropy motion planning,'' \emph{The International
  Journal of Robotics Research}, vol.~31, no.~7, pp. 855--871, 2012.

\bibitem{zhang2014applications}
W.~Zhang, H.~Wang, C.~Hartmann, M.~Weber, and C.~Schute, ``Applications of the
  cross-entropy method to importance sampling and optimal control of
  diffusions,'' \emph{SIAM Journal on Scientific Computing}, vol.~36, no.~6,
  pp. A2654--A2672, 2014.

\bibitem{testouri2023towards}
M.~Testouri, G.~Elghazaly, and R.~Frank, ``Towards a safe real-time motion
  planning framework for autonomous driving systems: {A}n {MPPI} approach,''
  \emph{arXiv preprint arXiv:2308.01654}, 2023.

\bibitem{mohamed2022autonomous}
I.~S. Mohamed, K.~Yin, and L.~Liu, ``Autonomous navigation of {AGVs} in unknown
  cluttered environments: {log-MPPI} control strategy,'' \emph{IEEE Robotics
  and Automation Letters}, vol.~7, no.~4, pp. 10\,240--10\,247, 2022.

\bibitem{ha2019topology}
J.-S. Ha, S.-S. Park, and H.-L. Choi, ``Topology-guided path integral approach
  for stochastic optimal control in cluttered environment,'' \emph{Robotics and
  Autonomous Systems}, vol. 113, pp. 81--93, 2019.

\bibitem{mohamed2020model}
I.~S. Mohamed, G.~Allibert, and P.~Martinet, ``Model predictive path integral
  control framework for partially observable navigation: {A} quadrotor case
  study,'' in \emph{2020 16th International Conference on Control, Automation,
  Robotics and Vision (ICARCV)}.\hskip 1em plus 0.5em minus 0.4em\relax IEEE,
  2020, pp. 196--203.

\bibitem{pravitra2021flying}
J.~Pravitra, E.~Theodorou, and E.~N. Johnson, ``Flying complex maneuvers with
  model predictive path integral control,'' in \emph{AIAA Scitech 2021 Forum},
  2021, p. 1957.

\bibitem{houghton2022path}
M.~D. Houghton, A.~B. Oshin, M.~J. Acheson, E.~A. Theodorou, and I.~M. Gregory,
  ``Path planning: Differential dynamic programming and model predictive path
  integral control on {VTOL} aircraft,'' in \emph{AIAA SCITECH 2022 Forum},
  2022, p. 0624.

\bibitem{higgins2023model}
J.~Higgins, N.~Mohammad, and N.~Bezzo, ``A model predictive path integral
  method for fast, proactive, and uncertainty-aware {UAV} planning in cluttered
  environments,'' \emph{arXiv preprint arXiv:2308.00914}, 2023.

\bibitem{nicolay2023enhancing}
P.~Nicolay, Y.~Petillot, M.~Marfeychuk, S.~Wang, and I.~Carlucho, ``Enhancing
  {AUV} autonomy with model predictive path integral control,'' \emph{arXiv
  preprint arXiv:2308.05547}, 2023.

\bibitem{hou2022robotic}
L.~Hou, H.~Wang, H.~Zou, and Y.~Zhou, ``Robotic manipulation planning for
  automatic peeling of glass substrate based on online learning model
  predictive path integral,'' \emph{Sensors}, vol.~22, no.~3, p. 1292, 2022.

\bibitem{yamamoto2020path}
K.~Yamamoto, R.~Ariizumi, T.~Hayakawa, and F.~Matsuno, ``Path integral policy
  improvement with population adaptation,'' \emph{IEEE Transactions on
  Cybernetics}, vol.~52, no.~1, pp. 312--322, 2020.

\bibitem{mohamed2021sampling}
I.~S. Mohamed, G.~Allibert, and P.~Martinet, ``Sampling-based {MPC} for
  constrained vision based control,'' in \emph{2021 IEEE/RSJ International
  Conference on Intelligent Robots and Systems (IROS)}.\hskip 1em plus 0.5em
  minus 0.4em\relax IEEE, 2021, pp. 3753--3758.

\bibitem{mohamed2021mppi}
I.~S. Mohamed, ``{MPPI-VS}: {S}ampling-based model predictive control strategy
  for constrained image-based and position-based visual servoing,'' \emph{arXiv
  preprint arXiv:2104.04925}, 2021.

\bibitem{costanzo2023modeling}
M.~Costanzo, G.~De~Maria, C.~Natale, and A.~Russo, ``Modeling and control of
  sampled-data image-based visual servoing with three-dimensional features,''
  \emph{IEEE Transactions on Control Systems Technology}, 2023, (Early Access).

\bibitem{macenski2023desks}
S.~Macenski, T.~Moore, D.~V. Lu, A.~Merzlyakov, and M.~Ferguson, ``From the
  desks of {ROS} maintainers: {A} survey of modern \& capable mobile robotics
  algorithms in the robot operating system 2,'' \emph{Robotics and Autonomous
  Systems}, p. 104493, 2023.

\bibitem{macenski2022robot}
S.~Macenski, T.~Foote, B.~Gerkey, C.~Lalancette, and W.~Woodall, ``Robot
  operating system 2: Design, architecture, and uses in the wild,''
  \emph{Science Robotics}, vol.~7, no.~66, p. eabm6074, 2022.

\bibitem{van2008graphical}
B.~Van Den~Broek, W.~Wiegerinck, and B.~Kappen, ``Graphical model inference in
  optimal control of stochastic multi-agent systems,'' \emph{Journal of
  Artificial Intelligence Research}, vol.~32, pp. 95--122, 2008.

\bibitem{thijssen2016path}
S.~A. Thijssen, ``Path integral control,'' Ph.D. dissertation, Radboud
  University, 2016.

\bibitem{gomez2016real}
V.~G{\'o}mez, S.~Thijssen, A.~Symington, S.~Hailes, and H.~J. Kappen,
  ``Real-time stochastic optimal control for multi-agent quadrotor systems,''
  in \emph{Proceedings of the Twenty-Sixth International Conference on
  Automated Planning and Scheduling (ICAPS)}, 2016, pp. 468--476.

\bibitem{wan2021cooperative}
N.~Wan, A.~Gahlawat, N.~Hovakimyan, E.~A. Theodorou, and P.~G. Voulgaris,
  ``Cooperative path integral control for stochastic multi-agent systems,'' in
  \emph{American Control Conference (ACC)}.\hskip 1em plus 0.5em minus
  0.4em\relax IEEE, 2021, pp. 1262--1267.

\bibitem{varnai2022multi}
P.~Varnai and D.~V. Dimarogonas, ``Multi-agent stochastic control using path
  integral policy improvement,'' in \emph{American Control Conference
  (ACC)}.\hskip 1em plus 0.5em minus 0.4em\relax IEEE, 2022, pp. 3406--3411.

\bibitem{pourchot2018cem}
A.~Pourchot and O.~Sigaud, ``{CEM-RL}: {C}ombining evolutionary and
  gradient-based methods for policy search,'' \emph{arXiv preprint
  arXiv:1810.01222}, 2018.

\bibitem{wen2018constrained}
M.~Wen and U.~Topcu, ``Constrained cross-entropy method for safe reinforcement
  learning,'' \emph{Advances in Neural Information Processing Systems},
  vol.~31, 2018.

\bibitem{amos2020differentiable}
B.~Amos and D.~Yarats, ``The differentiable cross-entropy method,'' in
  \emph{International Conference on Machine Learning}.\hskip 1em plus 0.5em
  minus 0.4em\relax PMLR, 2020, pp. 291--302.

\bibitem{zhang2022simple}
Z.~Zhang, J.~Jin, M.~Jagersand, J.~Luo, and D.~Schuurmans, ``A simple
  decentralized cross-entropy method,'' \emph{Advances in Neural Information
  Processing Systems}, vol.~35, pp. 36\,495--36\,506, 2022.

\bibitem{balci2022constrained}
I.~M. Balci, E.~Bakolas, B.~Vlahov, and E.~A. Theodorou, ``Constrained
  covariance steering based {Tube-MPPI},'' in \emph{2022 American Control
  Conference (ACC)}.\hskip 1em plus 0.5em minus 0.4em\relax IEEE, 2022, pp.
  4197--4202.

\bibitem{stulp2012path}
F.~Stulp and O.~Sigaud, ``Path integral policy improvement with covariance
  matrix adaptation,'' \emph{arXiv preprint arXiv:1206.4621}, 2012.

\bibitem{stulp2012policy}
------, ``Policy improvement methods: {B}etween black-box optimization and
  episodic reinforcement learning,'' \emph{HAL Open Science}, 2012.

\bibitem{lefebvre2019path}
T.~Lefebvre and G.~Crevecoeur, ``Path integral policy improvement with
  differential dynamic programming,'' in \emph{2019 IEEE/ASME International
  Conference on Advanced Intelligent Mechatronics (AIM)}.\hskip 1em plus 0.5em
  minus 0.4em\relax IEEE, 2019, pp. 739--745.

\bibitem{varnai2020path}
\BIBentryALTinterwordspacing
P.~Varnai and D.~V. Dimarogonas, ``Path integral policy improvement: {A}n
  information-geometric optimization approach,'' 2020. [Online]. Available:
  \url{10.13140/RG.2.2.13969.76645.}
\BIBentrySTDinterwordspacing

\bibitem{fu2020compound}
J.~Fu, C.~Li, X.~Teng, F.~Luo, and B.~Li, ``Compound heuristic information
  guided policy improvement for robot motor skill acquisition,'' \emph{Applied
  Sciences}, vol.~10, no.~15, p. 5346, 2020.

\bibitem{ba2021critic}
H.~Ba, J.~Fan, X.~Guo, and J.~Hao, ``Critic {PI2}: {M}aster continuous planning
  via policy improvement with path integrals and deep actor-critic
  reinforcement learning,'' in \emph{2021 6th IEEE International Conference on
  Advanced Robotics and Mechatronics (ICARM)}.\hskip 1em plus 0.5em minus
  0.4em\relax IEEE, 2021, pp. 716--722.

\bibitem{fleming2006controlled}
W.~H. Fleming and H.~M. Soner, \emph{Controlled Markov processes and viscosity
  solutions}.\hskip 1em plus 0.5em minus 0.4em\relax New York: Springer Science
  \& Business Media, 2006, vol.~25.

\bibitem{oksendal2013stochastic}
B.~Oksendal, \emph{Stochastic Differential Equations: An Introduction with
  Applications}.\hskip 1em plus 0.5em minus 0.4em\relax New York: Springer
  Science \& Business Media, 2013.

\bibitem{fleming2012deterministic}
W.~H. Fleming and R.~W. Rishel, \emph{Deterministic and Dtochastic Optimal
  Control}.\hskip 1em plus 0.5em minus 0.4em\relax New York: Springer Science
  \& Business Media, 2012, vol.~1.

\bibitem{kappen2007introduction}
H.~J. Kappen, ``An introduction to stochastic control theory, path integrals
  and reinforcement learning,'' in \emph{AIP Conference Proceedings}, vol. 887,
  no.~1.\hskip 1em plus 0.5em minus 0.4em\relax American Institute of Physics,
  2007, pp. 149--181.

\bibitem{kappen2011optimal}
H.~Kappen, ``Optimal control theory and the linear bellman equation,''
  \emph{Barber, D.; Cemgil, AT; Chiappa, S.(ed.), Bayesian time series models},
  pp. 363--387, 2011.

\bibitem{theodorou2011iterative}
E.~A. Theodorou, ``Iterative path integral stochastic optimal control: Theory
  and applications to motor control,'' Ph.D. dissertation, University of
  Southern California, 2011.

\bibitem{thijssen2015path}
S.~Thijssen and H.~Kappen, ``Path integral control and state-dependent
  feedback,'' \emph{Physical Review E}, vol.~91, no.~3, p. 032104, 2015.

\bibitem{fleming1995risk}
W.~H. Fleming and W.~M. McEneaney, ``Risk-sensitive control on an infinite time
  horizon,'' \emph{SIAM Journal on Control and Optimization}, vol.~33, no.~6,
  pp. 1881--1915, 1995.

\bibitem{theodorou2015nonlinear}
E.~A. Theodorou, ``Nonlinear stochastic control and information theoretic
  dualities: {C}onnections, interdependencies and thermodynamic
  interpretations,'' \emph{Entropy}, vol.~17, no.~5, pp. 3352--3375, 2015.

\bibitem{liptser1977statistics}
R.~S. Liptser and A.~N. Shiriaev, \emph{Statistics of Random Processes I:
  General Theory}, 2nd~ed.\hskip 1em plus 0.5em minus 0.4em\relax Heidelberg:
  Springer-Verlag Berlin, 2001.

\bibitem{liptser2013statistics}
------, \emph{Statistics of Random Processes II: Applications}, 2nd~ed.\hskip
  1em plus 0.5em minus 0.4em\relax Heidelberg: Springer-Verlag Berlin, 2001.

\bibitem{kupcsik2013data}
A.~Kupcsik, M.~Deisenroth, J.~Peters, and G.~Neumann, ``Data-efficient
  contextual policy search for robot movement skills,'' in \emph{Proceedings of
  the National Conference on Artificial Intelligence (AAAI)}.\hskip 1em plus
  0.5em minus 0.4em\relax Bellevue, 2013.

\bibitem{hansen2001completely}
N.~Hansen and A.~Ostermeier, ``Completely derandomized self-adaptation in
  evolution strategies,'' \emph{Evolutionary Computation}, vol.~9, no.~2, pp.
  159--195, 2001.

\bibitem{hansen2016cma}
N.~Hansen, ``The {CMA} evolution strategy: A tutorial,'' \emph{arXiv preprint
  arXiv:1604.00772}, 2016.

\bibitem{deisenroth2013survey}
M.~P. Deisenroth, G.~Neumann, J.~Peters \emph{et~al.}, ``A survey on policy
  search for robotics,'' \emph{Foundations and Trends{\textregistered} in
  Robotics}, vol.~2, no. 1--2, pp. 1--142, 2013.

\bibitem{vinogradska2020quadrature}
J.~Vinogradska, B.~Bischoff, J.~Achterhold, T.~Koller, and J.~Peters,
  ``Numerical quadrature for probabilistic policy search,'' \emph{IEEE
  Transactions on Pattern Analysis and Machine Intelligence}, vol.~42, no.~1,
  pp. 164--175, 2020.

\bibitem{deisenroth2013gaussian}
M.~P. Deisenroth, D.~Fox, and C.~E. Rasmussen, ``Gaussian processes for
  data-efficient learning in robotics and control,'' \emph{IEEE Transactions on
  Pattern Analysis and Machine Intelligence}, vol.~37, no.~2, pp. 408--423,
  2013.

\bibitem{thor2020generic}
M.~Thor, T.~Kulvicius, and P.~Manoonpong, ``Generic neural locomotion control
  framework for legged robots,'' \emph{IEEE Transactions on Neural Networks and
  Learning Systems}, vol.~32, no.~9, pp. 4013--4025, 2020.

\bibitem{schaal2005learning}
S.~Schaal, J.~Peters, J.~Nakanishi, and A.~Ijspeert, ``Learning movement
  primitives,'' in \emph{Robotics research. the eleventh international
  symposium}.\hskip 1em plus 0.5em minus 0.4em\relax Springer, 2005, pp.
  561--572.

\bibitem{ijspeert2013dynamical}
A.~J. Ijspeert, J.~Nakanishi, H.~Hoffmann, P.~Pastor, and S.~Schaal,
  ``Dynamical movement primitives: {L}earning attractor models for motor
  behaviors,'' \emph{Neural computation}, vol.~25, no.~2, pp. 328--373, 2013.

\bibitem{levine2016end}
S.~Levine, C.~Finn, T.~Darrell, and P.~Abbeel, ``End-to-end training of deep
  visuomotor policies,'' \emph{The Journal of Machine Learning Research},
  vol.~17, no.~1, pp. 1334--1373, 2016.

\bibitem{polydoros2017survey}
A.~S. Polydoros and L.~Nalpantidis, ``Survey of model-based reinforcement
  learning: Applications on robotics,'' \emph{Journal of Intelligent \& Robotic
  Systems}, vol.~86, no.~2, pp. 153--173, 2017.

\bibitem{garaffa2021reinforcement}
L.~C. Garaffa, M.~Basso, A.~A. Konzen, and E.~P. de~Freitas, ``Reinforcement
  learning for mobile robotics exploration: {A} survey,'' \emph{IEEE
  Transactions on Neural Networks and Learning Systems}, vol.~34, no.~8, pp.
  3796--3810, 2021.

\bibitem{moerland2023model}
T.~M. Moerland, J.~Broekens, A.~Plaat, C.~M. Jonker \emph{et~al.},
  ``Model-based reinforcement learning: {A} survey,'' \emph{Foundations and
  Trends{\textregistered} in Machine Learning}, vol.~16, no.~1, pp. 1--118,
  2023.

\bibitem{sutton1990integrated}
R.~S. Sutton, ``Integrated architectures for learning, planning, and reacting
  based on approximating dynamic programming,'' in \emph{Machine Learning
  Proceedings 1990}.\hskip 1em plus 0.5em minus 0.4em\relax Elsevier, 1990, pp.
  216--224.

\bibitem{deisenroth2011pilco}
M.~Deisenroth and C.~E. Rasmussen, ``{PILCO}: {A} model-based and
  data-efficient approach to policy search,'' in \emph{Proceedings of the 28th
  International Conference on machine learning (ICML-11)}, 2011, pp. 465--472.

\bibitem{janner2019trust}
M.~Janner, J.~Fu, M.~Zhang, and S.~Levine, ``When to trust your model:
  {M}odel-based policy optimization,'' \emph{Advances in neural information
  processing systems}, vol.~32, pp. 1--9, 2019.

\bibitem{yu2020mopo}
T.~Yu, G.~Thomas, L.~Yu, S.~Ermon, J.~Y. Zou, S.~Levine, C.~Finn, and T.~Ma,
  ``{MOPO}: {M}odel-based offline policy optimization,'' \emph{Advances in
  Neural Information Processing Systems}, vol.~33, pp. 14\,129--14\,142, 2020.

\bibitem{kotecha2003gaussian}
J.~H. Kotecha and P.~M. Djuric, ``Gaussian sum particle filtering,'' \emph{IEEE
  Transactions on signal processing}, vol.~51, no.~10, pp. 2602--2612, 2003.

\bibitem{liu2001monte}
J.~S. Liu, \emph{{M}onte {C}arlo Strategies in Scientific Computing}.\hskip 1em
  plus 0.5em minus 0.4em\relax New York: Springer, 2001, vol.~75.

\bibitem{zhang2021path}
Q.~Zhang and Y.~Chen, ``Path integral sampler: {A} stochastic control approach
  for sampling,'' in \emph{International Conference on Learning
  Representations}, 2022.

\bibitem{patil2023risk}
A.~Patil, Y.~Zhou, D.~Fridovich-Keil, and T.~Tanaka, ``Risk-minimizing
  two-player zero-sum stochastic differential game via path integral control,''
  \emph{arXiv preprint arXiv:2308.11546}, 2023.

\bibitem{wan2021distributed}
N.~Wan, A.~Gahlawat, N.~Hovakimyan, E.~A. Theodorou, and P.~G. Voulgaris,
  ``Distributed algorithms for linearly-solvable optimal control in networked
  multi-agent systems,'' \emph{arXiv preprint arXiv:2102.09104}, 2021.

\bibitem{song2022generalization}
L.~Song, N.~Wan, A.~Gahlawat, C.~Tao, N.~Hovakimyan, and E.~A. Theodorou,
  ``Generalization of safe optimal control actions on networked multiagent
  systems,'' \emph{IEEE Transactions on Control of Network Systems}, vol.~10,
  no.~1, pp. 491--502, 2022.

\bibitem{song2023safety}
L.~Song, P.~Zhao, N.~Wan, and N.~Hovakimyan, ``Safety embedded stochastic
  optimal control of networked multi-agent systems via barrier states,'' in
  \emph{2023 American Control Conference (ACC)}.\hskip 1em plus 0.5em minus
  0.4em\relax IEEE, 2023, pp. 2554--2559.

\bibitem{watterson2018trajectory}
M.~Watterson, S.~Liu, K.~Sun, T.~Smith, and V.~Kumar, ``Trajectory optimization
  on manifolds with applications to {SO}(3) and $\mathbb{R}^3 \times
  \mathbb{S}^{2}$,'' in \emph{Robotics: Science and Systems}, 2018, p.~9.

\bibitem{bonalli2019trajectory}
R.~Bonalli, A.~Bylard, A.~Cauligi, T.~Lew, and M.~Pavone, ``Trajectory
  optimization on manifolds: {A} theoretically-guaranteed embedded sequential
  convex programming approach,'' \emph{arXiv preprint arXiv:1905.07654}, 2019.

\bibitem{watterson2020trajectory}
M.~Watterson, S.~Liu, K.~Sun, T.~Smith, and V.~Kumar, ``Trajectory optimization
  on manifolds with applications to quadrotor systems,'' \emph{The
  International Journal of Robotics Research}, vol.~39, no. 2-3, pp. 303--320,
  2020.

\bibitem{osa2022motion}
T.~Osa, ``Motion planning by learning the solution manifold in trajectory
  optimization,'' \emph{The International Journal of Robotics Research},
  vol.~41, no.~3, pp. 281--311, 2022.

\bibitem{boumal2014manopt}
N.~Boumal, B.~Mishra, P.-A. Absil, and R.~Sepulchre, ``Manopt, a {M}atlab
  toolbox for optimization on manifolds,'' \emph{The Journal of Machine
  Learning Research}, vol.~15, no.~1, pp. 1455--1459, 2014.

\bibitem{boumal2023introduction}
N.~Boumal, \emph{An introduction to optimization on smooth manifolds}.\hskip
  1em plus 0.5em minus 0.4em\relax Cambridge University Press, 2023.

\bibitem{menegaz2018unscented}
H.~M. Menegaz, J.~Y. Ishihara, and H.~T. Kussaba, ``Unscented {K}alman filters
  for {R}iemannian state-space systems,'' \emph{IEEE Transactions on Automatic
  Control}, vol.~64, no.~4, pp. 1487--1502, 2018.

\bibitem{brossard2020code}
M.~Brossard, A.~Barrau, and S.~Bonnabel, ``A code for unscented {K}alman
  filtering on manifolds {(UKF-M)},'' in \emph{IEEE International Conference on
  Robotics and Automation (ICRA)}.\hskip 1em plus 0.5em minus 0.4em\relax IEEE,
  2020, pp. 5701--5708.

\bibitem{li2020unscented}
K.~Li, F.~Pfaff, and U.~D. Hanebeck, ``Unscented dual quaternion particle
  filter for {SE}(3) estimation,'' \emph{IEEE Control Systems Letters}, vol.~5,
  no.~2, pp. 647--652, 2020.

\bibitem{cantelobre2020real}
T.~Cantelobre, C.~Chahbazian, A.~Croux, and S.~Bonnabel, ``A real-time
  unscented {K}alman filter on manifolds for challenging {AUV} navigation,'' in
  \emph{IEEE/RSJ International Conference on Intelligent Robots and Systems
  (IROS)}.\hskip 1em plus 0.5em minus 0.4em\relax IEEE, 2020, pp. 2309--2316.

\end{thebibliography}

\end{document}